\documentclass[11pt]{article}

\newif\ifSciRepSubmission % use composite eps instead of separate pdf figs
\ifSciRepSubmission
\else
  \pdfoutput=1 % pdflatex compiler
\fi

\usepackage{fullpage}

\newif\ifdraft % False by default?
\newif\ifCompileSI
\CompileSItrue

\ifdraft
   \usepackage[left,modulo]{lineno}
   \usepackage{graphicx,eso-pic,xcolor}
   
   \makeatletter
   \AddToShipoutPicture{%
   \setlength{\@tempdimb}{.5\paperwidth}%
   \setlength{\@tempdimc}{.5\paperheight}%
   \setlength{\unitlength}{1pt}%
   \put(\strip@pt\@tempdimb,\strip@pt\@tempdimc){%
       \makebox(-500,200){\rotatebox{90}{\textcolor[gray]{0.70}%
          {\Large \textsf{Draft of \today}}}}
     }%
   }
   \makeatother
\fi

\ifdraft
  \usepackage[normalem]{ulem} 
  \newcommand{\rednote}[1]{\textcolor{red}{#1}}
  \newcommand{\redfootnote}[1]{\footnote{\textcolor{red}{#1}}}
\else  
  \newcommand{\sout}[1]{\iffalse #1 \fi}
  \newcommand{\rednote}[1]{\iffalse #1 \fi}
  \newcommand{\redfootnote}[1]{\iffalse #1 \fi}
\fi

\usepackage[font={small}]{caption}

\usepackage{cite} % http://tex.stackexchange.com/a/15114/29550
\usepackage{graphicx,xcolor}
\definecolor{linkColor}{rgb}{0.6471, 0.1098, 0.1882} % Crimson
\definecolor{urlColor}{rgb}{0., 0., 1.} % Blue
\usepackage[colorlinks=true,citecolor=linkColor,linkcolor=linkColor,urlcolor=urlColor,breaklinks=true]{hyperref}

\newcommand{\refMethods}{\hyperref[sec:methods]{Methods section}}

\newcommand{\SInfo}{Supplementary Information} %% TODO
\newcommand{\SItem}{Supplementary} %% TODO

\usepackage{amsmath}
\usepackage[binary-units=true]{siunitx} 
\DeclareSIUnit\ppm{ppm}
\DeclareSIUnit\pixel{px}
\DeclareSIUnit\inch{\si{\arcsecond}}

\newcommand{\curv}{k_-}
\newcommand{\curvdir}{X_-} %{\hat{\mathbf{e}}_-}

\newcommand{\ulcase}{\lowercase} 
\newcommand{\subfig}[1]{\ulcase{#1}}
\newcommand{\subcap}[1]{(\textbf{\ulcase{#1}})} 
\newcommand{\capheader}[1]{\textbf{#1.}}

\usepackage{graphicx}

\graphicspath{ {figures/}
               {figures/algorithm_outline/}
               {figures/complexity/}
               {figures/snaps1/}
               {figures/snaps3/}
               {figures/supporting_figs/}
               {figures/supporting_figs/EA_vs_EDCircles/}
               {figures/composite/}
             }

\newcommand{\complexity}[2]{#1_sliced_img_#2}

\newcommand{\videoDemo}[2]{demo_sliced_img_#2}

\newcommand{\outlinea}{bone_rescaled_unprocessed_sliced_fig_40}
\newcommand{\outlineb}{directed_ridges_over_principal_curv_sliced_fig_40}
\newcommand{\outlinec}{RidgeDirectedCircleHoughTransform}
\newcommand{\outlined}{rings_subpxl_unprocessed_sliced_fig_40}
\newcommand{\outlinee}{masked_ridges_over_unprocessed_sliced_fig_40}
\newcommand{\outlinef}{RidgeDirectedCircleHoughTransform_threshed_smoothed_normed}

\newcommand{\CalibExmpl}{sugar_over_glass_example_tracers_based_calibration}
\newcommand{\CalibCurv}{sugar_over_glass_25_tracers_inverted_quadratic_calibration}

\newcommand{\FullFrame}{unprocessed_sliced_fig_46}
\newcommand{\trajectoriesExample}{isometric_parallel_proj_first_12s_colour_is_global_speed_reduced}

\newcommand{\EDComparable}{EAvsEDC_sliced_img_3445316_679}
\newcommand{\EAoutperformsa}{EAvsEDC_sliced_img_3445265_249}
\newcommand{\EAoutperformsb}{EAvsEDC_sliced_img_3445290_964}
\newcommand{\EAoutperformsc}{EAvsEDC_sliced_img_3445293_821}

\usepackage{paralist}

\usepackage{tikz}
\usepackage{pgfplots}
\usetikzlibrary{positioning}
\usetikzlibrary{calc} 

\newcommand{\tikzLabel}[2]{
    \begin{tikzpicture}[font=\sffamily]%\bfseries]%Helvetika] 
        \node[anchor=south west,inner sep=0] (image) at (0,0){#2};
            \begin{scope}[x={(image.south east)},y={(image.north west)}]
                \draw (2.5mm,0.93) node {#1};
            \end{scope}
        \end{tikzpicture}
    }
\newcommand{\tikzLabelSnaps}[2]{
    \begin{tikzpicture}[font=\sffamily]%\bfseries]%Helvetika] 
        \node[anchor=south west,inner sep=0] (image) at (0,0){#2};
            \begin{scope}[x={(image.south east)},y={(image.north west)}]
                \draw (2mm,0.87) node {#1};
            \end{scope}
        \end{tikzpicture}
    }

\newlength\snaptotalwidth
\definecolor{lightblue}{rgb}{0.145,0.6666,1}
\newlength{\FSZ}
\newcommand{\drawvideo}[3]{% [0 0.25 0.5 0.75 1 1.25 1.5]
   \noindent\pgfmathsetlength{\FSZ}{\snaptotalwidth/#2}
   \begin{tikzpicture}[outer sep=0pt,inner sep=0pt,x=\FSZ,y=\FSZ]
   \draw[color=lightblue!50!black] (0,0) node[outer sep=0pt,inner sep=0pt,text
       width=\snaptotalwidth,minimum height=0] (video) {\noindent#3};
   \path [fill=lightblue!50!black,line width=0pt] 
     (video.north west) rectangle ([yshift=\FSZ] video.north east) 
    \foreach \x in {1,2,...,#2} {
      {[rounded corners=0.6] ($(video.north west)+(-0.7,0.8)+(\x,0)$) rectangle +(0.4,-0.6)}
    }
;
   \path [fill=lightblue!50!black,line width=0pt] 
     ([yshift=-1\FSZ] video.south west) rectangle (video.south east) 
    \foreach \x in {1,2,...,#2} {
      {[rounded corners=0.6] ($(video.south west)+(-0.7,-0.2)+(\x,0)$) rectangle +(0.4,-0.6)}
    }
;
   \foreach \x in {1,...,#1} {
     \draw[color=lightblue!50!black] ([xshift=\x\snaptotalwidth/#1] video.north
     west) -- ([xshift=\x\snaptotalwidth/#1] video.south west);
   }
   \foreach \x in {0,#1} {
     \draw[color=lightblue!50!black] ([xshift=\x\snaptotalwidth/#1,yshift=1\FSZ]
     video.north west) -- ([xshift=\x\snaptotalwidth/#1,yshift=-1\FSZ] video.south west);
   }
   \end{tikzpicture}
}

\begin{document}

%%%%%%%%%%%%%%%%%% title page information %%%%%%%%%%%%%%%%%%
\title{Robust and highly performant ring detection algorithm 
        for 3d particle tracking using 2d microscope imaging}

\author{Eldad Afik\footnote{To whom correspondence should be addressed; 
  email: \href{mailto:eldada.afik@weizmann.ac.il}{eldad.afik@weizmann.ac.il}}\\
   \normalsize{Department of Physics of Complex Systems, Weizmann Institute of Science,}\\
   \normalsize{Rehovot 76100, Israel}\\
}
\date{}

\maketitle

\ifdraft
%\else
  \begin{linenumbers}
\fi

\begin{abstract}
  Three-dimensional particle tracking is an essential tool in studying dynamics
  under the microscope, namely, fluid dynamics in microfluidic devices,
  bacteria taxis, cellular trafficking.
  The 3d position can be determined using 2d imaging alone by measuring the
  diffraction rings generated by an out-of-focus fluorescent particle, imaged
  on a single camera.
  Here I present a ring detection algorithm exhibiting a high detection rate,
  which is robust to the challenges arising from ring occlusion, inclusions and
  overlaps, and allows resolving particles even when near to each other.
  It is capable of real time analysis thanks to its high performance and low
  memory footprint.
  The proposed algorithm, an offspring of the circle Hough transform, 
  addresses the need to efficiently trace the trajectories of many particles
  concurrently, when their number in not necessarily fixed, by solving a
  classification problem, and overcomes the challenges of finding
  local maxima in the complex parameter space which results from ring clusters and
  noise.
  Several algorithmic concepts introduced here can be advantageous
  in other cases, particularly when dealing with noisy and sparse data.
  The implementation is based on open-source and cross-platform software
  packages only, making it easy to distribute and modify.
  It is implemented in a microfluidic experiment allowing real-time
  multi-particle tracking at \SI{70}{\Hz}, achieving a detection rate which exceeds
  94\% and only 1\% false-detection.
\end{abstract}

%%condensed abstract (500 chars):
% Three-dimensional particle tracking is an essential tool in studying dynamics
% under the microscope. In this work, the 3d position of a fluorescent particle
% is determined using out-of-focus 2d imaging alone, using a single camera. I
% present a ring detection algorithm, which is robust to the challenges arising
% from particles vicinity. It is implemented in a microfluidic experiment
% allowing real-time multi-particle tracking at 70Hz, achieving a 94% detection
% rate and only 1% false-detection.

%%%%%%%%%%%%%%%%%%%%%%%%%%  main body  %%%%%%%%%%%%%%%%%%%%%%%%%%
%\section{Introduction}
The study of dynamics often relies on
tracking objects under the microscope. Indeed, precise and robust particle
tracking is essential in many fields, including studies of micro-Rheology
\cite{Crocker2000,Lau2003}, chaotic dissipative flows
\cite{Burghelea2004,Gerashchenko2005}, feedback for micro-manipulation
\cite{Gosse2002}, and other soft condensed matter physics and engineering problems.
Moreover, microfluidic systems play a growing role as part of lab-on-a-chip
apparatus in micro-chemistry \cite{deMello2006,McMullen2010}, bioanalytics
\cite{Khandurina2000,Zhang2006}, and other bio-medical research and engineering
applications \cite{Sackmann2014}. 
Yet, detailed characterisation of the flow and transport phenomena at the
micro-scale is still a non-trivial task. In general, the motion is three
dimensional and automated tracking is cumbersome from the perspective of both
instrumentation and software.
Setting up several viewing angles as done for large systems
\cite{Bourgoin2006} becomes even more complicated in
microscopic systems, while scanning through the third axis, e.g. confocal
microscopy, clearly compromises temporal resolution and concurrency. 
In this work the three-dimensional positions of fluorescent particles are inferred from
the information encoded in the diffraction rings which result from
out-of-focus imaging, converting the 3d localisation problem to an image
analysis problem of ring detection. 
\iffalse
In the process of tracking particles in a microfluidic experiment, 
I was confronted with the need to detect and localise
particles in 3d, based on 0.7 megapixel images, acquired at \SI{70}{\Hz}, containing
roughly 50 particles per frame.
However, my needs are not unique because 
\fi

The development of the method presented here was motivated by the study of pair
dispersion in a chaotic flow \cite{Afik2014}, taking place in a microfluidic
tube of \SI{140}{\um} -- the observation volume is larger by more than three orders of
magnitude with respect to those reported in Refs.~\cite{Gosse2002} and
\cite{Speidel2003}.
The experiments consist of tracking tracers advected by the flow, at seeding
levels of several tens to hundreds in the observation window, where it is
necessary that the particles are resolved even when nearby to each other. The
typical flow rates dictate sampling rates of \SI{70}{\Hz} whereas the statistical
nature of the problem requires data acquisition over weeks.
Using a standard epi-fluorescence microscope and the fact that the parameters
of the most visible ring can be mapped to the 3d position of the tracer, the
particle localisation problem is converted to a circle detection problem.
The constraints set by the nature of the experiment require an image analysis
algorithm that is robust not only to the noise of the image acquisition
process, but to rings overlaps, inclusions and occlusions as well. The typical
complexity of the images is exemplified in a sub-frame from our experiment
presented in \autoref{fig:video_snapshots}\subfig{a}. 
In addition, the data flow is of about
\SI[per-mode=symbol]{180}{\giga\byte\per\hour}, an overwhelming rate which
demands the optimisation of the algorithm for real-time analysis.

In this presentation I will focus on the development of an algorithm for this
purpose. The key steps for achieving high-performance are introduced following the
presentation of the main concepts which contribute to the robustness of the
algorithm.
The application for particle tracking is presented and discussed in
the \SInfo; further technical details of the optical apparatus can be found in
the \refMethods .

Imagine for the moment that you have successfully identified which of the
pixels in the image reside on a ring. The issue of doing so will be addressed
later on. 
Given this set of coordinates, it may seem straightforward to find the
parameters of the circles which best fit them. However there is a missing piece
of information here, that is, which sub-sets of pixels belong together to form
a ring.
Moreover, we do not know a priori how many rings there are in the image and
there may be false detected coordinates which we would like to disregard.
Therefore, we need some method to classify/cluster the coordinates into
sub-sets, each sub-set matching a single ring.

A circle in a two-dimensional image is uniquely specified by three parameters.
In this work two designate the centre of the circle and the third
specifies its radius. The parameter space of all possible circles is therefore
three-dimensional. One can detect circles in an image by mapping
the image intensity field to the circle parameter space. Peaks in this parameter
space imply a circle well represented in the image. 
One approach to achieve this mapping is via a discrete Radon transform, which
for the purposes of this presentation translates to convolving the image with a
mask of a ring \cite{Ginkel2004}. Since each candidate radius calls for a
separate convolution, this results in visiting all the pixels in the image over
and over. Recall that the outer-most ring is sufficient for 3d localisation of
the imaged particles. Hence it is worth noting that even at moderate rings
densities, the pixels lying on the outer-most ones consist a small fraction of
the pixel population, less than 2\%  in my case.
When there are more than a couple of potential radii this procedure would
perform a plethora of useless computations \cite{Ginkel2004}. This fact
directs to another approach, which may seem equivalent yet explicitly exploits
this information sparsity  -- the circle Hough transform \cite{Duda1972}:
each pixel votes for all the candidate points in the parameter space of which
it may be part. In this way, every pixel is visited once,
potentially reducing the computation time by orders of magnitude. The discrete
version of the parameter space is commonly referred to as the array of
accumulators. During the voting procedure each vote increments an accumulator
by one.
Alas, in the literature of Computer Vision and Pattern Recognition it is well
known that the standard circle Hough transform is rather demanding both for
large memory requirements, which grow with the radii range, as well as for
its 3d nature which renders peak finding in the parameter space a difficult
task to tackle \cite{Yuen1990,Huang2012}. 

One way to address these computational challenges is to resort to lower
dimensionality circle Hough transforms, but these usually miss circles having
nearby centres and are less robust, resulting in higher false positive and
false negative errors rates \cite{Yuen1990}. 
Another path is to randomly sub-sample the information content in the image, giving way to
non-deterministic methods; see Ref.~\cite{Huang2012} and references therein. 
However, due to their random nature these methods suffer from inferior detection rates and
accuracy when compared with the deterministic ones \cite{Huang2012}.

For these reasons I developed an algorithm which is an
offspring of the full 3d-circle Hough transform, yet the local maxima detection
issues are addressed and it shows high performance and a small memory
footprint.

%%%%%%%%%%%%%%%%%%%%%%%%

%%%   Results   %%%
\section*{Results}%: The proposed algorithm and empirical evaluation}
\paragraph*{\iffalse algorithm rationale introduction \fi} 

\ifSciRepSubmission
\else
\begin{figure*}[<+htpb+>]
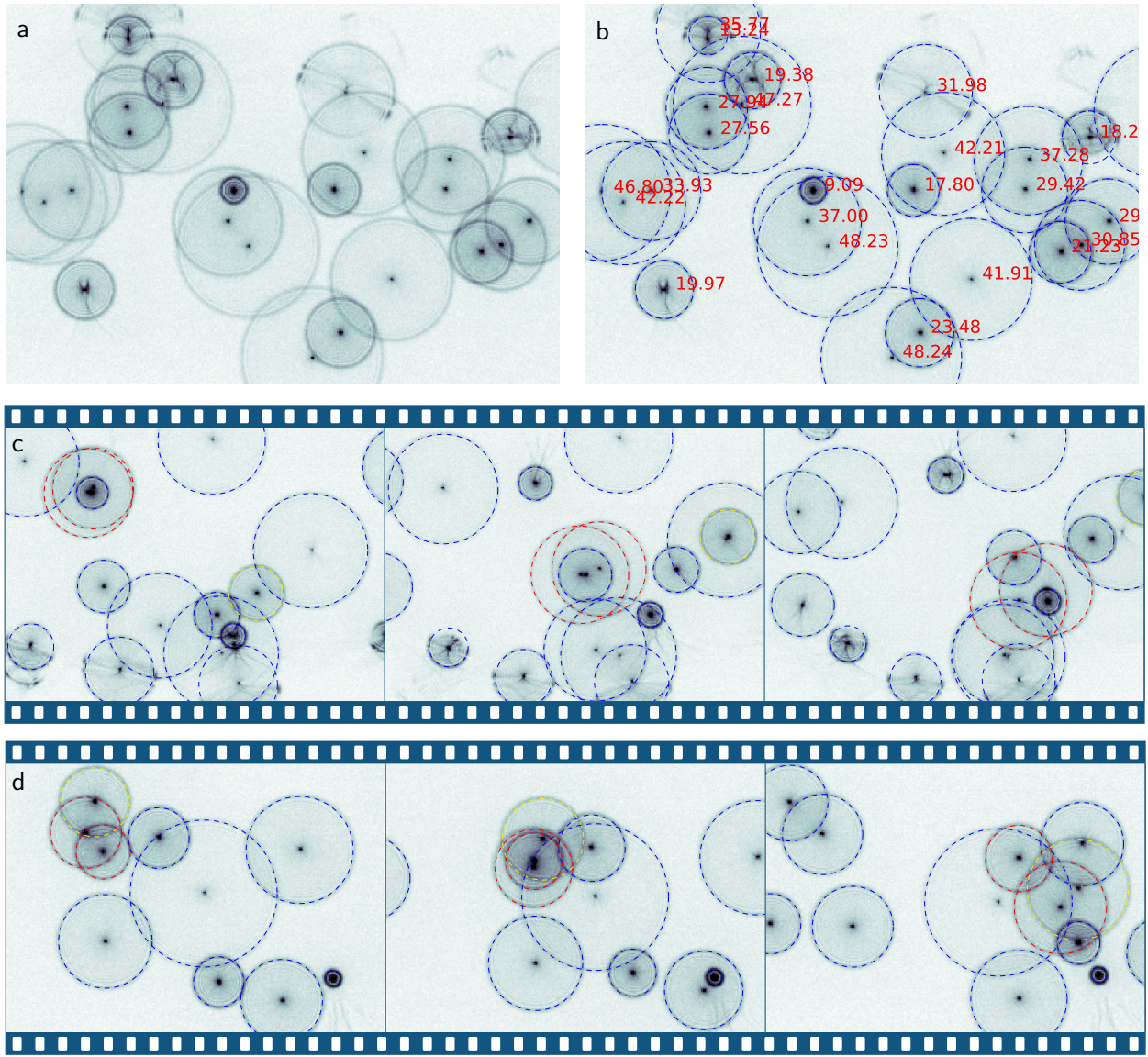

    \begin{center}
      \centerline{
        \tikzLabel{a}{%
            \includegraphics[width=0.485\snaptotalwidth]{\complexity{raw}{3445224_105}}}
        \tikzLabel{b}{%
            \includegraphics[width=0.485\snaptotalwidth]{\complexity{demo}{3445224_105}}}
          }\vspace{-2mm}
        \newlength\snapwidth
        \setlength\snapwidth{0.3333\snaptotalwidth}%
        \tikzLabelSnaps{c}{
        \drawvideo{3}{50}{%
            \includegraphics[width=\snapwidth]{\videoDemo{1}{3445227_562}}%
            \includegraphics[width=\snapwidth]{\videoDemo{1}{3445227_848}}%
            \includegraphics[width=\snapwidth]{\videoDemo{1}{3445227_991}}%
        }}%
        
        \vspace{2mm}
        
        \tikzLabelSnaps{d}{%
        \drawvideo{3}{50}{%
            \includegraphics[width=\snapwidth]{\videoDemo{3}{3445409_866}}%
            \includegraphics[width=\snapwidth]{\videoDemo{3}{3445409_995}}%
            \includegraphics[width=\snapwidth]{\videoDemo{3}{3445410_252}}%
        }}
    \end{center}
    \caption{\capheader{Snapshots from the experiment and a demonstration of the algorithm
    robustness}
        \subcap{a} typical image complexity is exemplified in an unprocessed
        sub-frame consisting of 1/9 part of the full frame, corresponding to
        lateral dimension of \SI{215}{\um}$\times$\SI{315}{\um}. The axial range available
        for the particles is \SI{140}{\um}.
        \subcap{b} the corresponding analysis result; in red are the radii in
        pixels units.
        \subcap{c} \& \subcap{d} time sequences of sub-frames (\SI{400}{\ms} each). 
        Red coloured particles in \subcap{c} demonstrate pair dispersion, in
        which the algorithm is required to resolve rings with similar
        parameters.
        The yellow particle in \subcap{d} shows radius change corresponding to a
        downwards translation.
        Each sub-frame in \subcap{c} \& \subcap{d} images a box which lateral
        dimensions is \SI{190}{\um}$\times$\SI{270}{\um}.
        }\label{fig:video_snapshots}
\end{figure*}
\fi

The key steps of the algorithm are conceptually outlined as follows:
\begin{inparaenum}[(i)]
    \item detect directed ridges;
    \item map the directed ridges to the parameter space of circles;
    \item detect local maxima via radius dependent smoothing and normalisation;
    \item classify the coordinates of the ridge pixels according to the peaks
        in the circle parameter space, and fit each sub-set to a circle,
        achieving sub-pixel accuracy.
\end{inparaenum}
This outline is presented graphically in \autoref{fig:algorithm_outline} for a
small sub-frame containing two fluorescent particles.

\ifSciRepSubmission
\else
\begin{figure*}[<+htpb+>]
    \begin{center} 
    \newlength\outlineSize
    \setlength\outlineSize{0.185\textheight}
    \tikzLabel{a}{%
        \includegraphics[height=\outlineSize]{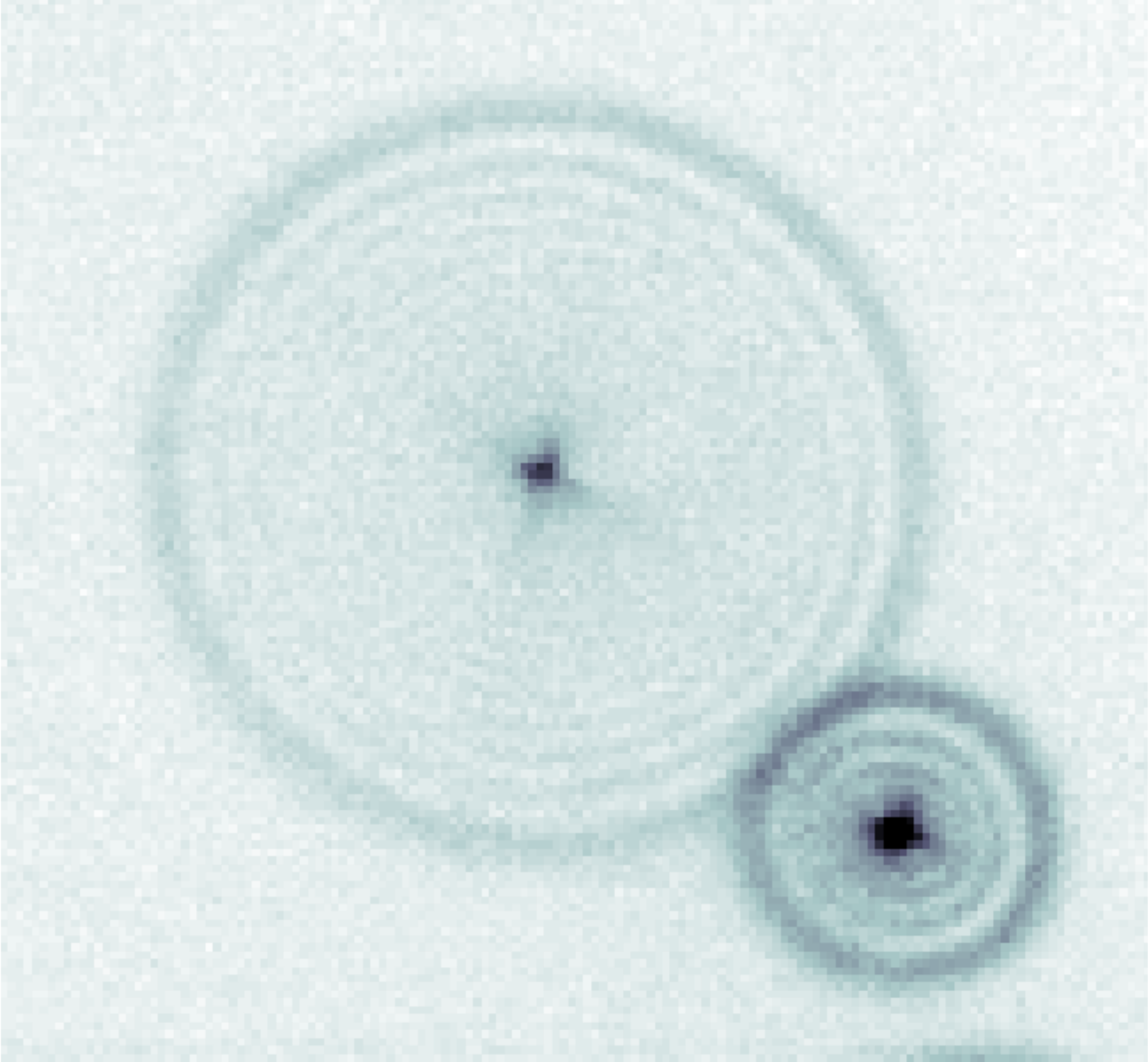}}\hspace{1mm}
    \tikzLabel{b}{%
        \includegraphics[height=\outlineSize]{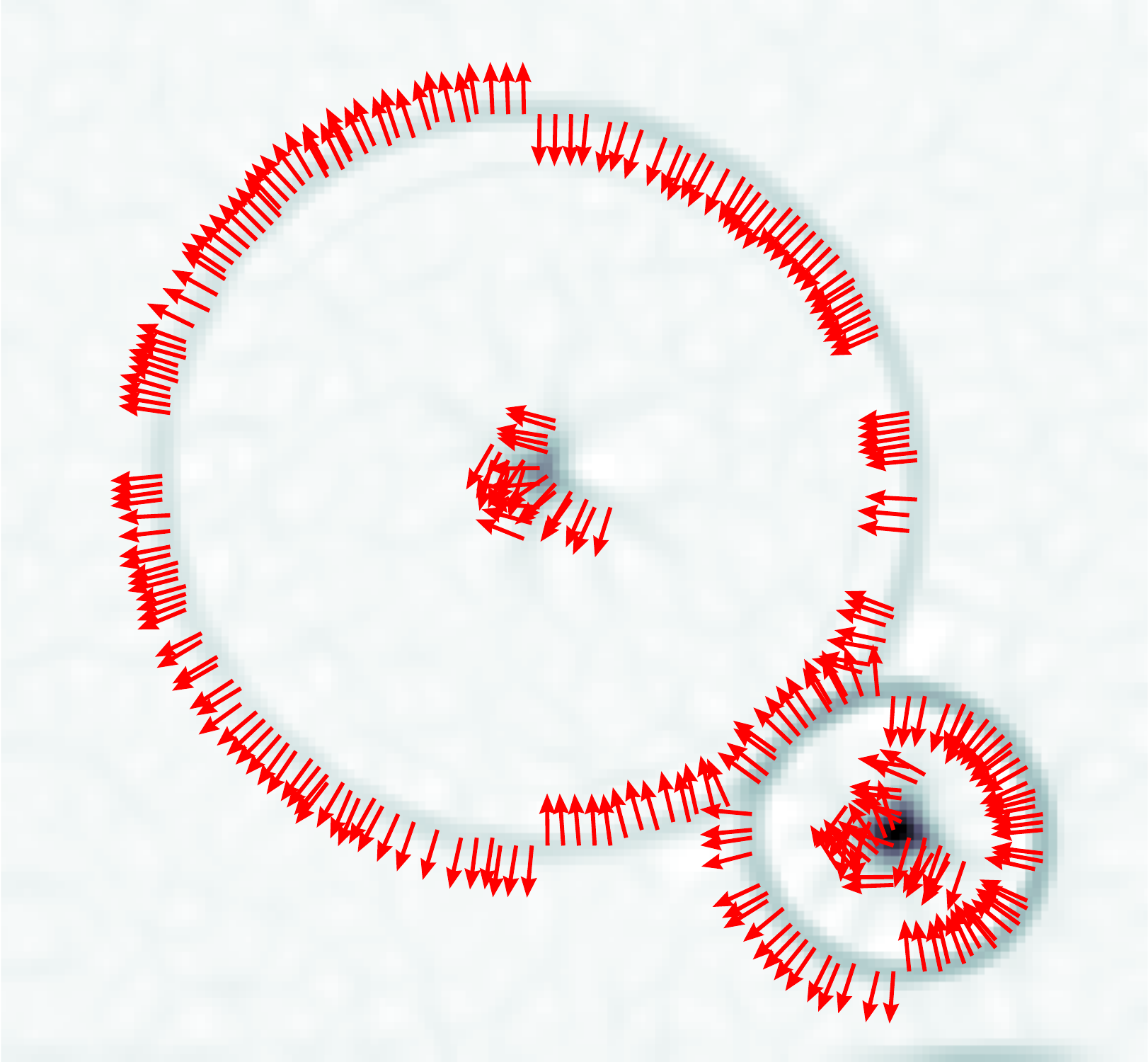}}\hspace{1mm}
    \raisebox{-0.25\outlineSize}{\tikzLabel{c}{%
        \includegraphics[height=\outlineSize]{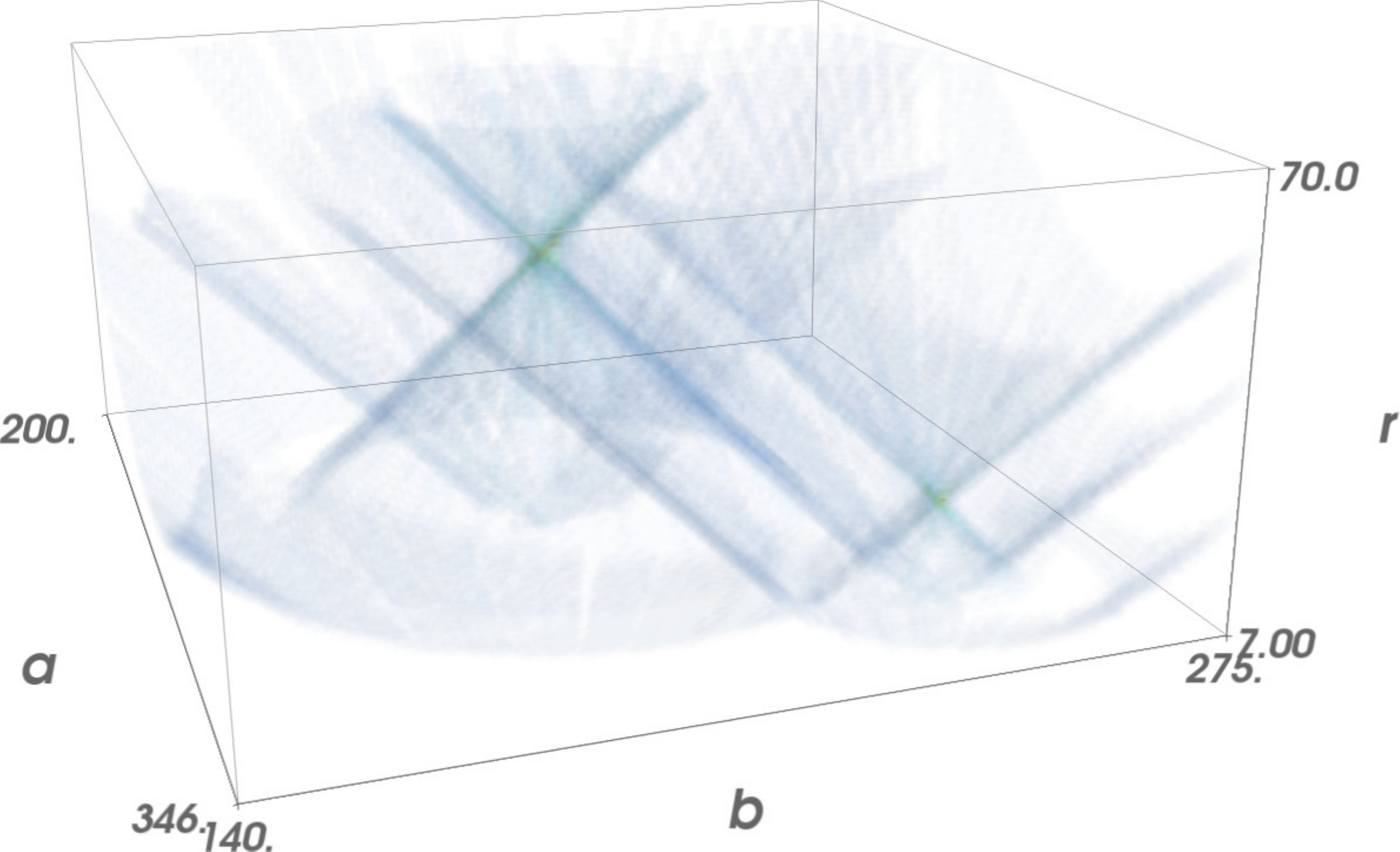}}}

    \tikzLabel{f}{%
        \includegraphics[height=\outlineSize]{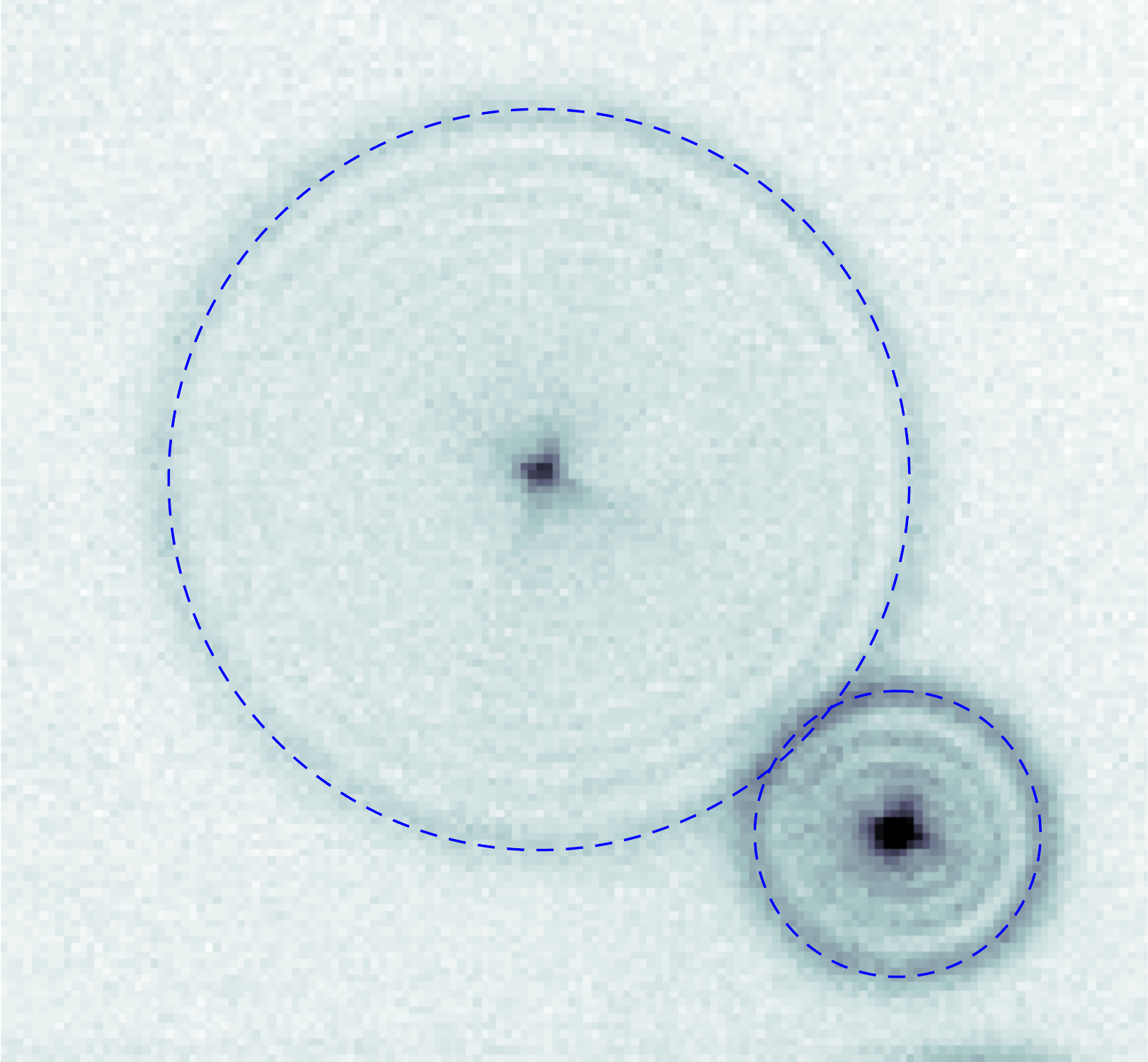}}\hspace{1mm}
    \tikzLabel{e}{%
        \includegraphics[height=\outlineSize]{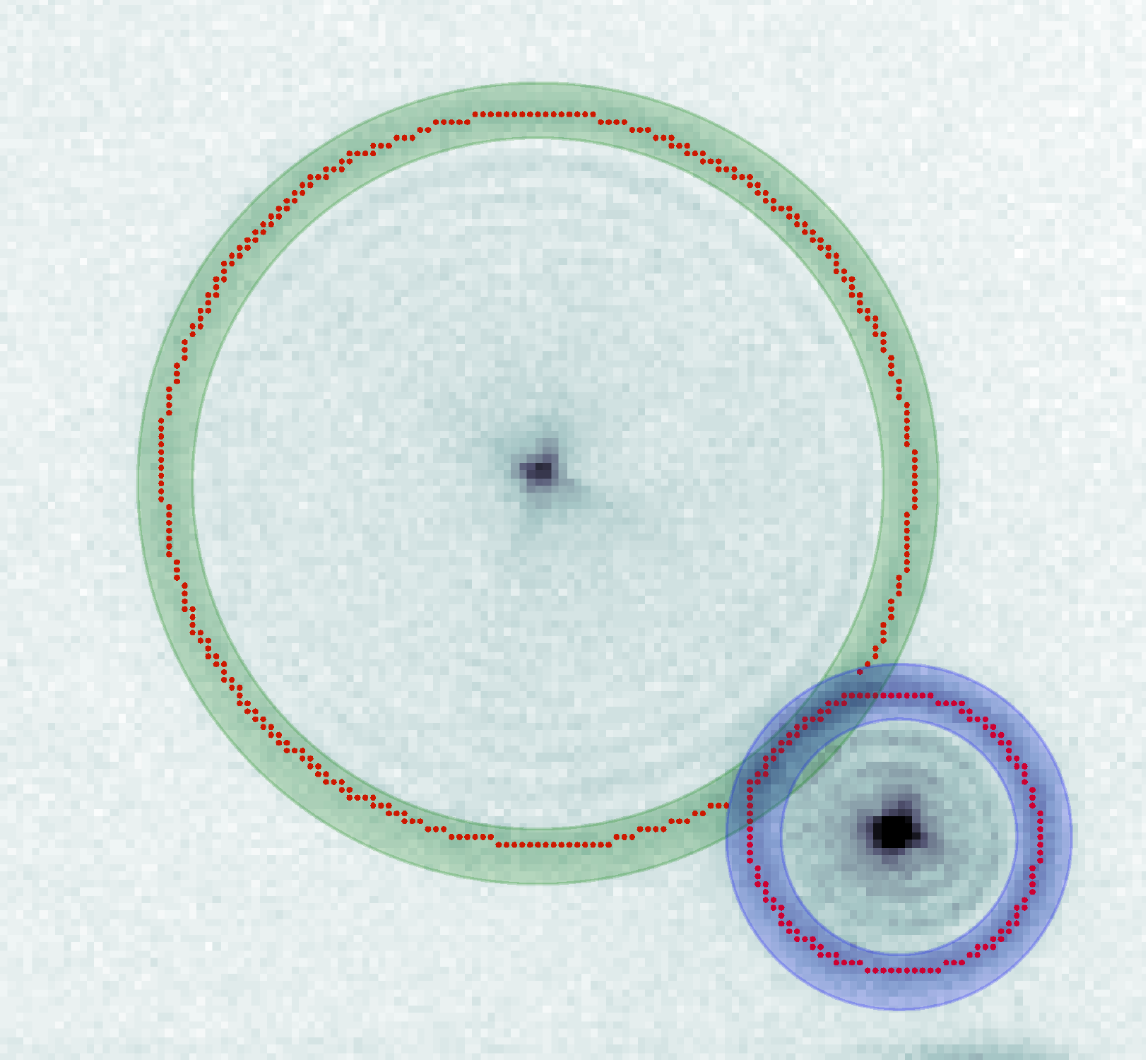}}\hspace{1mm}
    \raisebox{0.25\outlineSize}{\tikzLabel{d}{%
        \includegraphics[height=\outlineSize]{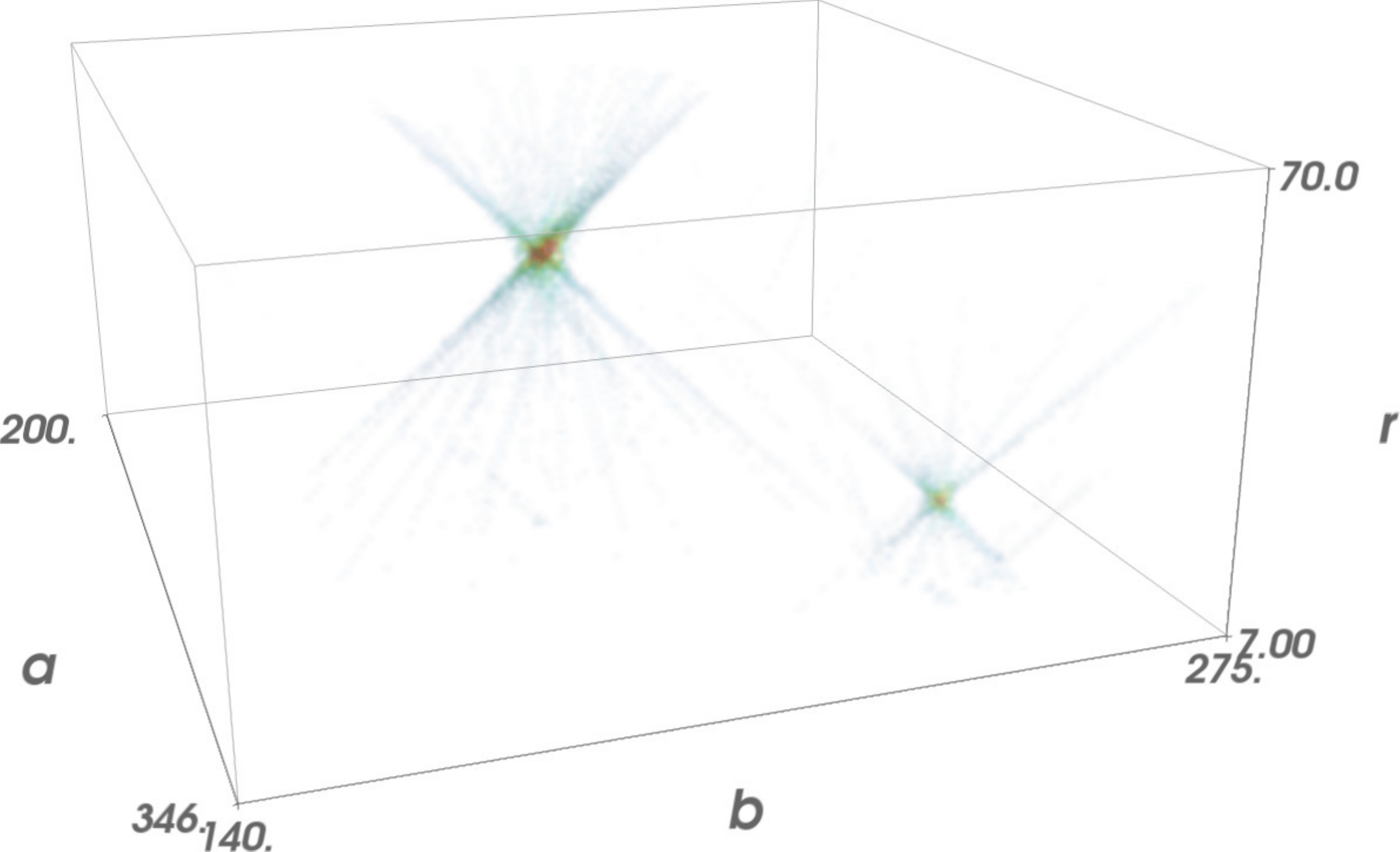}}}
   \end{center}
   \caption{\capheader{Algorithm outline}
        \subcap{a} raw sub-image containing two fluorescent particles; note
        that the inner rings of each particle are thinner than the outer most
        one. This scale separation admits suppression of all but the outer most
        ring via Gaussian smoothing (to ease visualisation the contrast was
        enhanced in the images on the expense of the central peak of the
        diffraction pattern);
        \subcap{b} ridge detection: the ridges are defined using a differential
        geometric descriptor and shown here as arrows representing $\curvdir$,
        the principal direction, corresponding to $\curv$, the least principal
        curvature, which is plotted in the background. The arrows originate
        from the ridge pixel. Note that the inner rings are successfully
        suppressed based on the scale separation. To ease visualisation every
        second detected ridge is omitted;
        \subcap{c} circle Hough transform: directed ridges $\to$ circle
        parameter space;
        \subcap{d} local maxima detection: radius dependent smoothing of the
        parameter space as well as normalisation by 1/r and thresholding
        greatly emphasise the local maxima representing the rings in the image;
        \subcap{e} sub-pixel accuracy: based on the detected rings, annulus
        masks (blue and green annuli in the figure) allow classification of
        ridge pixels (red points) and sub-pixel accuracy is achieved via circle
        fitting. Note the discarded directed ridges of the central peak (in
        \subcap{b}) as they do not belong to any local maxima in the processed
        circle parameter space \subcap{d};
        \subcap{f} the output: best fit circle for the ridge pixels of the
        outer-most ring of each particle.
        } \label{fig:algorithm_outline}
\end{figure*}
\fi

\paragraph*{Directed ridge detection \& votes collection}

The first step is locating the pixels of interest. Like many other feature
detection algorithms, the standard circle Hough transform relies on an edge
detection step, where edges are the borders of dark and bright regions.
As the images contain rings rather then filled circles, I chose to
implement an algorithm that detects ridges, thin curves which are brighter than
their neighbourhood, rather than edges. This exhibits better consistency. 
First note that the ring of interest is thicker than the inner ones. This is
advantageous as the image admits scale selection \cite{Lindeberg1999} -- the
inner rings can be suppressed using a Gaussian smoothing having the appropriate
scale, approximately that of the most visible ring.
Ridges are then found using a differential geometric descriptor
\cite{Lindeberg1999}, which defines ridge pixels using the following two
properties:
\begin{inparaenum}[(i)]
    \item Negative least principal curvature, $\curv < 0$; 
    \item $\curv$ is a local minimum along the direction of the associated
        principal direction $\curvdir$. 
\end{inparaenum}
The principal curvatures are the eigenvalues of the Hessian, the matrix of the
image second spatial derivatives. Here $\curv$ and $\curvdir$ denote the
smaller principal curvature and the corresponding eigenvector.
This is demonstrated for the two fluorescent particles in 
\autoref{fig:algorithm_outline}\subfig{a}, where this sub-image is analysed for directed
ridges, represented by arrows overlaid on the $\curv$ field as background of
\autoref{fig:algorithm_outline}\subfig{b}.
Note that $\curvdir$ is collinear with the direction to the centre of the ring.
I use this to significantly reduce the complexity of the voting procedure, in a
similar way to the gradient directed circle Hough transform \cite{Kimme1975}
-- each directed ridge pixel votes for all candidate points in the 3d parameter
space of circles, provided that the circle centre is within
the $r_{min}$ to $r_{max}$ range, directed along $\curvdir$.
%\begin{algorithm}[H]
%    \DontPrintSemicolon
%    \For{each $(x,y) \in ridge$}{ 
%        \For{each $r \in [r_{min},r_{max}]$}{
%            $(a,b) = (x,y) \pm r \cdot \curvdir $\; 
%            HoughSpace$(r,a,b) \ +\!= \ 1 $ 
%            } 
%        } 
%\end{algorithm} 

\paragraph*{Local maxima detection in a noisy parameter space: radius dependent
smoothing \& normalisation}

As votes from the ridge pixels accumulate, each ring in the image transforms
into two mirroring coaxial cones, aligned along the radius axis, having a joint
apex. This procedure results in a discrete scalar function over a 3d box.
This is demonstrated in \autoref{fig:algorithm_outline}\subfig{c}. 
The coordinates of the apexes, which are the local maxima of this function, are
the candidate circle parameters.
In practice, there are many sources which render the resulting circle parameter
space very noisy: The raw image is a discrete representation of the
intensity field, the image acquisition process itself is not noiseless, and the
image complexity mentioned above, all may result in errors in the detected
ridge position and direction, as well as false ridge detection and false
negatives.
Note that the deviation from ideal voting due to the error in determining the
ridge direction grows linearly with the radius. Therefore, each equi-radius
level of the ring parameter space is smoothed using Gaussian weights, whose
width is proportional to the radius.
Next, note that finding local maxima in the 3d parameter space requires the
comparison of accumulators in different equi-radius levels, which asks for some
normalisation as larger rings are expected to receive more votes. This leads to
a natural normalisation by $1/r$, following which an ideal ring is expected
to receive $2\pi$ votes. \autoref{fig:algorithm_outline}\subfig{d} shows how this
procedure simplifies the parameter space. 
Local maxima can then be located by nearest-neighbours comparison.

\paragraph*{Ridge points classification \& sub-pixel accuracy}
The local maxima identified in the parameter space induce a classification on
the ridge coordinates: for each peak an annulus mask is formed and a best
fitting circle is found for all ridge coordinates within the annulus. Ridge
pixels which are not covered by any annulus mask are not fitted for. This is
desired as these usually result from non-circular features in the image or
noise. See the example in \autoref{fig:algorithm_outline}\subfig{e}. In this way
sub-pixel precision is achieved.

\paragraph*{Empirical detection and error rates}

Examining the robustness of the algorithm on the experimental data reveals a
detection rate that exceeds 94\% with only 1\% false-detection; for further details
see Robustness assessment in the \refMethods.

To demonstrate the excellence of these results let us compare the robustness
with a recently published algorithm --- the EDCircles algorithm introduced in
Ref.~\cite{Akinlar2013}. Founded on the mathematical theory of perception
\cite{Desolneux2007} it detects contiguous edge segments
and employs the Helmholtz principle for controlling false detections.
It was chosen as the competitor for this examination for two main reasons: 
\begin{inparaenum}[(i)]
  \item it is parameter-free and so the comparison is insensitive to the choice
    of input parameters; and
  \item the results presented in the manuscript \cite{Akinlar2013} are very
    promising --
     the EDCircles was demonstrated to exhibit a much better detection rate
    when compared with the state-of-the-art lower dimensionality Circle Hough Transform implemented
    in OpenCV \cite{OpenCV}, referred to as 21HT in Ref. \cite{Yuen1990}.
\end{inparaenum}

In practice, the EDCircles showed a detection rate lower than 61\% and nearly 2\%
false-detection.
While the EDCircles detected 21\% of the rings missed by the algorithm proposed
here, the latter detected successfully more than 88\% of those missed by its
competitor for this comparison; examples can be found in \SItem~%
\ifCompileSI
  \autoref{fig:EDCircles_comparison}%
\else
  Fig.~S3% 
\fi; further details of the test and results are summarised in Comparative
assessment of the algorithm robustness in the \refMethods~.

For precision and accuracy estimation, in particular in the light of particle
localisation, see Experimental details and Precision assessment in the
\refMethods~as well as \SItem~%
\ifCompileSI
  \autoref{fig:rad2z}
\else
  Fig.~S1 
\fi and the accompanying caption.

\paragraph*{Key algorithmic optimisations for 
memory requirement \& temporal performance}

As was mentioned above, it is desired for our purposes to have the images
processed in real-time. Here I briefly outline the key ideas behind the
optimisation of the algorithm, the full details are available in the
open-source code itself (see the \refMethods). 
%(\url{https://github.com/eldad-a/ridge-directed-ring-detector}). 
%\cite{Afik2014a}. 
The first key point for the algorithm optimisation is the splitting of the
voting procedure --  the votes are recorded for each ridge pixel as it is
detected, such that ridge detection and votes collection are done in
\emph{one-pass}.

The population of the parameter space is performed at a separate stage, which
leads to the second key point. Instead of holding an array for the full
parameter space, only two sub-spaces are maintained, consisting of three
consecutive equi-radius levels; the first for the raw parameter sub-space, the
second for the smoothed and normalised one, where local maxima are searched
for. The equi-radius levels are populated and processed one by one.
Only the accumulators exceeding an integer vote threshold are regarded as
hotspots and are mapped to the smoothed and normalised sub-space. 
Each time a radius-level is completed, regarded here as the ``top'' one, the
hotspots in the level beneath, the ``middle'' one, are verified to exceed a
pre-set floating point threshold, a fraction of $2\pi$. Those which do are
searched for local maxima by a nearest neighbour comparison within a
$3\times3\times3$ voxels box.
Once this search is completed, the ``bottom'' level is no longer needed and a
cyclic permutation takes place where the ``bottom'' level becomes the new
``top''.
This allows \emph{memory recycling} and avoids the need to initialise big arrays
of zeros to represent the full 3d parameter space.

\emph{Registering modified array elements and undoing} is the last key point.
The radius-levels are required to be blank prior to their population. Recalling
the sparsity of the parameter space (see
Figs.~\ref{fig:algorithm_outline}\subfig{c}~and~\ref{fig:algorithm_outline}\subfig{d}
for example), going over all its elements is a waste of processing time.
Instead, each time an array element is modified for the first time its indices
are registered. Once the search for peaks is done, all modifications to the
``bottom'' radius-level are undone, preparing it for reuse. This lifts the need
to clean the whole array. 

The combination of \emph{memory recycling} with \emph{registering modified
array elements and undoing} reduces the computation time 
and in my case results in a nearly ten times less memory consumption. 
In fact, the size of the arrays representing the parameter space kept in memory
is now fixed and no longer grows with the radii range.
For preliminary testing purposes, I first implemented the algorithm outlined in
the beginning of the Results section (as well as in
\autoref{fig:algorithm_outline}) using convolutions
and other array based operations. 
The final implementation, inspired by the circle Hough transform and including
the above optimisations, is more than 50 times faster.
This is attributable to the reduction in the number of operations required once
the sparsity of the data is taken advantage of.
Further details and explanations can be found in the \refMethods~and
the Detailed algorithm section of the \SInfo. %\autoref{sec:detailed_algorithm}.

%%%%%%%%%%%%%%%%%%%%%%%%

%%%   Discussion   %%%

\section*{Discussion}
In this work I have presented a new algorithm to analyse images of complex
annular patterns. Image complexity and noise often result in a challenging parameter
space where local maxima are difficult to find, a problem not addressed within
the classical Hough transform algorithm.
The main novelty introduced here to overcome this difficult task and to gain
robustness are the radius dependent smoothing and normalisation.
The resulting detection and error rates are very promising, even more so
in the light of alternative methods.
As it was already mentioned in the introduction the non-deterministic or
randomised methods typically provide a gain in the temporal performance but
suffer in reliability when it comes to detection and error rates
\cite{Huang2012}.
The EDCircles algorithm \cite{Akinlar2013} was chosen as a competitor for the
comparative assessment of robustness mainly as it was reported to outperform
the state-of-the-art implementation of the natural competitor ---
OpenCV's deterministic Circle Hough Transform \cite{OpenCV}.
The algorithm proposed here demonstrated a detection rate higher by more
than 50\% and a nearly three times smaller false reports rate.

Several algorithmic concepts have been introduced to improve memory
requirements and temporal performance, of highest importance are those referred
above as \emph{registering modified array elements and undoing} as well as
\emph{memory recycling}. These have been empirically shown to reduce memory
consumption by nearly ten times and result in an over fifty times faster
analysis rate. 
These can be advantageous for other algorithms as well, particularly when the
data is sparse.

Though the development of this algorithm was motivated by the analysis of fluorescence
microscopy images, it is more general and can be applied to other cases as
well. The interpretation of the Hough transform as a classification/clustering
algorithm has a wider potential than merely image analysis. To name a physical
example is the case of particle jets emerging from several sources of unknown
loci. A dataset consisting of the positions and momenta of the particles at a
certain time is analogous to that of the directed ridges and the jet sources
can be identified with the local maxima over the parameter space.

The method I introduced above is currently implemented in an experiment which
requires long unsupervised measurements lasting for days at high temporal
resolution, sampling a volume of interest which contains tens to hundreds of
particles. Thanks to the fact that the whole volume of interest is sampled at
once by a single 2d image, concurrency is achieved.
The use of LED and the relatively short exposure times can be potentially
exploited to avoid photo-damage and bleaching. 
It is demonstrated to be robust to the overlap, inclusion and occlusion of the
ring pattern of the imaged particles. 
It features high performance admitting real-time applications.
A discussion of this approach for particle tracking in light of other methods
\cite{Cheong2010,Dixon2011,Kao1994,Huang2008,Babcock2012}
can be found in the second section
\iffalse\autoref{sec:application_for_tracking}\fi of the \SInfo.

This method paves the way for studies of 3d flows in microfluidic devices.
Its robustness to the vicinity of particles to each other allows to study the
dynamics of particle pairs \cite{Afik2014}, triples, etc. 
As such it also has a potential for biomedical research. A possible immediate 
application is the detailed characterisation of the transport induced by the
presence of cells in confined flows, a phenomenon presented in Ref.
\cite{Amini2012}.
Together with the development of high signal tracers, labelling techniques and
sensitive cameras, this method may be useful in other life sciences studies
such as cellular trafficking \cite{Brandenburg2007}, cell migration
\cite{Friedl2000} and bacterial taxis \cite{Xie2011}.

%%%%%%%%%%%%%%%%%%%%%%%%

%%%   Methods   %%%
{\section*{Methods}\label{sec:methods}}
\paragraph*{Algorithm implementation}
%\subsection*{Algorithm implementation}
I implemented this algorithm relying on freely available open-source and
cross-platform software packages only. The source is available online 
(\url{https://github.com/eldad-a/ridge-directed-ring-detector}). 
%\cite{Afik2014a} .
Most of the heavy lifting is achieved using the Cython language \cite{Cython}.
It has a Python-like syntax from which a C code is automatically generated and
compiled. This allows the code to be short and easy to read while enjoying
the performance of C.  
For example, this implementation exploits the Numpy/Cython strided direct data
access \cite{Cython,SciPy} by fully sorting the votes.
In the image pre-processing step the image is smoothed using a Gaussian
convolution and the smoothed image spatial derivatives are calculated using a
5$\times$5 2\textsuperscript{nd} order Sobel operator; these operations are
done using OpenCV's Python bindings \cite{OpenCV}.

The equi-radius levels of the circle parameter space are smoothed using
Gaussian weights $ \exp\left\{
    -\frac{1}{2}\left[\frac{x-x_{hotspot}}{\sigma}\right]^2\right\}
$, whose width is proportional to the radius $\sigma(r)\propto r$. The
explicit form, $\sigma(r) = 0.05r + 0.25$, was found empirically. The slope
coefficient is interpreted as accounting for $\sim0.1$rad uncertainty in the
direction of $\curvdir$.

\paragraph*{Experimental details}
The imaging system consists of an inverted fluorescence microscope (IMT-2,
Olympus), mounted with a Plan-Apochromat 20$\times$/0.8NA objective (Carl
Zeiss) and a fluorescence filter cube; a Royal-Blue LED (Luxeonstar) served
for the fluorophore excitation.  A CCD (GX1920, Allied Vision Technologies) was
mounted via zoom and 0.1$\times$ c-mount adapters (Vario-Orthomate 543513 and
543431, Leitz), sampling at \SI{70}{\Hz}, \SI{968}{\pixel}$\times$\SI{728}{\pixel}, 
covering \SI{810}{\um}$\times$\SI{610}{\um} laterally.

The experiments were conducted in a microfluidic device, implemented in
polydimethylsiloxane elastomer by soft lithography, consisting of a curvilinear
tube (see grey broken line in \SItem~%
\ifCompileSI
  \autoref{fig:2d->3d}%
\else
  Fig.~S2%
\fi\subfig{a}). 
The rectangular cross-section of the tube was measured to be of \SI{140}{\um} depth and
\SI{185}{\um} width.
The working fluid consisted of polyacrylamide in aqueous sugar (sucrose and
sorbitol) syrup, seeded with fluorescent particles (1 micron 15702
Fluoresbrite\textsuperscript{\textregistered} YG Carboxylate particles,
PolySciences Inc.).
The flow was driven by gravity.

For the empirical calibration, the same working fluid was
sandwiched between two microscope glass slides. The separation distance
between the slides was set to \SI{161}{\um} by micro-spheres (4316A PS NIST certified
calibration and traceability, Duke Standards) serving as spacers.
The microscope objective was translated in steps of \SI{2}{\um} to acquire images
of the tracers at different off-focus distances $\Delta z$. The microscope
focus knob was manipulated by a computer controlling a stepper motor.
During the rest stages of the objective, the ring radii of every
detected tracer were averaged over 210 frames spanning \SI{3}{\second}. 
Due to the high viscosity of the fluid, 1100 times larger than water viscosity, tracer
motion due to diffusion is negligible during this time interval.
The median of the estimated standard-deviations of the data presented in
\SItem~%
\ifCompileSI
  \autoref{fig:rad2z}%
\else
  Fig.~S1%
\fi\subfig{a} is \SI{0.03}{\pixel} and the maximal is \SI{0.27}{\pixel}.
In practice, to account for the uncertainties in finding the focal position and
due to optical aberrations, 25 tracers dispersed throughout the observation
volume were accounted for.
Their curves were aligned via shifting $\Delta z$ by the larger root of a
quadratic polynomial fit. Then, the conversion function $r^{-1}(r)$
was obtained by inversion of the quadratic polynomial fit accounting for all
the data together; see \SItem~%
\ifCompileSI
  \autoref{fig:rad2z}%
\else
  Fig.~S1%
\fi\subfig{b}.
The resulting root-mean-squared-error, 
$\sqrt{\langle \left( \Delta z - r^{-1}(r) \right)^2 \rangle} =$~\SI{1.97}{\um},
and the maximal measured absolute error is \SI{5.35}{\um}; these reflect the
uncertainty due to the empirical calibration procedure taken here.
Finally, the out-of-focus distance of the objective $\Delta z$ has to be
converted to the physical distance via multiplication by the ratio of the
refractive indices, 1.58 in this case. The observed axial range exceeds
\SI{180}{\um}.

\paragraph*{Robustness assessment}
In order to estimate the robustness of the algorithm, images from the
chaotic flow experiment were analysed and cropped to a sub-region of
\SI{340}{\pixel}$\times$\SI{370}{\pixel}. The analysis results of 600 such
sub-frames were examined.
These sub-frames contained 14.3 rings on average, out of which 67.7\% were in
ring clusters (overlap and inclusion configurations).  
This examination shows an average of 6.8\% False-Negative errors. In some
sub-frames a ghost ring would appear accompanied by a strong distortion of the
tracer image in its real position. This is attributed the microfluidic walls
and observed only when a particle is very close to the wall. Excluding from the
statistics two such tracers, the False-Negative error rate is reduced,
corresponding to a detection rate of 94.7\%.
This examination does not show any significant sensitivity to rings overlap and
inclusion. On the contrary, 95.5\% of the rings in clusters were identified
correctly while only 0.8\% of the reported rings in clusters were non-existing
particles. 

\paragraph*{Comparative assessment of the algorithm robustness}
\iffalse
To illustrate the high-quality of the above results,
\SItem~%
\ifCompileSI
  \autoref{fig:2d->3d}%
\else
  Fig.~S2%
\fi\subfig{a} was analysed by the on-line demo of the
parameter-free EDCircles algorithm, provided by the authors of
Ref.~\cite{Akinlar2013}. 
The results are provided here for comparison.
The EDCircles demo detected 35 circles and 3 ellipses. 
One of the ellipses is a false merging of two rings, which were
correctly resolved by the algorithm presented here. Excluding these two,
the EDCircles did not detect 19 out of the other 56 rings found by the
method proposed in this manuscript.
None of the particles correctly reported by the former was missed by
the latter.
\fi
To demonstrate the high-quality of the above results, a comparison was made
against the on-line demo of the EDCircles, provided by the authors of
Ref.~\cite{Akinlar2013}.
The tests were conducted on a subset of 151 images taken from the same experiment as
in the Robustness assessment above, cropped to the same sub-region of
\SI{340}{\pixel}$\times$\SI{370}{\pixel}. 
In this case the images were first cropped and exported to the PNG format prior to
the analysis, the format for which the EDCircles on-line demo exhibited the
best detection rate. In this comparison rings whose centre lay outside the
cropped image were not considered, as well as those whose visible perimeter was
less than a half of the complete one; the average ring number was found to be
13.4.
\iffalse
Ellipses reported by the EDCircles demo were counted as correctly detected as
long as they did not deviate too much from a ring in the image
\fi
Few examples are presented in \SItem~%
\ifCompileSI
  \autoref{fig:EDCircles_comparison}%
\else
  Fig.~S3% 
\fi . 

The EDCircles demo detected 60.9\% of the rings (False-Negative error
rate of 39.2\%); in contrast, the method proposed here showed a detection rate
of 94.3\%.
While 1.7\% of the reported detections by the EDCircles demo were
False-Positive, only 0.6\% of the rings reported by the new algorithm presented
here were non-existing or erroneous ones.
Out of those particles missed by the proposed method 21.1\% were detected by the
EDCircles demo; in contrast, 88.6\% of those missed by the competitor algorithm
were detected by the one proposed here.

\redfootnote{TODO: mention tests made by Yoav on synthetic gaussian rings,
accuracy and performance}

\paragraph*{Performance assessment}
The performance assessment is based the analysis of 1500 full frames containing
50 particles on average; for a typical example see
\SItem~%
\ifCompileSI
  \autoref{fig:2d->3d}%
\else
  Fig.~S2%
\fi\subfig{a}.
The test was run on an i7-3820 CPU desktop, running Ubuntu linux operating
system.
A single process analysed at an average rate of \SI{6.28}{\Hz}.
To achieve higher performance as required for our experiments, I use the
multiprocessing package of Python, exploiting the multi-core processors
available on modern computers. The analysis rate scales linearly with the
number of processes. No deterioration of the processing rate per core was
noted (tested up to twice the core number with hyper-threading). 
Based on a producer-consumer model, one can even transparently distribute the
workload among several computers if needed. This is partially attributable to
the small memory footprint of the algorithm.

\paragraph*{Precision assessment}
To estimate the precision of the presented localisation method,
smoothing splines were applied to the reconstructed trajectories providing an
estimator for the error variance $\sigma^2$; see Ref.~\cite{Wahba1983}. The mean
values are as follows: $\sigma_x =$~\SI{0.134}{\um}, $\sigma_y
=$~\SI{0.135}{\um} and $\sigma_z =$~\SI{0.434}{\um}, for $x$ and $y$ denoting the
lateral coordinates, and $z$ the axial one. This axial uncertainty corresponds
to 0.3\% of the axial range covered by the particles.
According to the data provided in Ref.~\cite{Gosse2002} a value
of 0.1\% was achieved and a similar one in Ref.~\cite{Speidel2003}.
The uncertainty estimates reported here account for noise which rise not only
due to the image analysis and the multi-particle scenario, but also due to
other sources, namely the motion of the particles and the linking procedure. 
The details are as follows.

From a 3m30s \iffalse 0:03:35.54746\fi measurement, 2014 
trajectories which span more than 1s were analysed (discarding shorter ones).
Particle positions, converted to microns, were linked to reconstruct their
trajectories; for a sub-sample see \SItem~%
\ifCompileSI
  \autoref{fig:2d->3d}%
\else
  Fig.~S2%
\fi\subfig{b}. 
The linking algorithm was adapted from the code accompanying
Ref.~\cite{Kelley2011}, generalised to n-dimensions, the kinematic model was
modified to account for accelerations, and a memory feature was introduced to
account for occasional misses.
Finally, a natural cubic smoothing spline was applied to smooth-out noise
and obtain an estimate of particle velocity and acceleration
\cite{Wasserm2007,Ahnert2007}.
The smoothing parameter was automatically set using Vapnik's measure, using a code
adapted from the Octave splines package \cite{Krakauer2014}.

%\section*{}
\ifdraft
%\else
  \end{linenumbers}
\fi

%%%%%%%%%%%%%%%%%%%%%%% References %%%%%%%%%%%%%%%%%%%%%%%%%

%\bibliographystyle{naturemag_no_url}
%\bibliography{/home/eldada/Copy/eBooks/EA_academic_references}

\begin{thebibliography}{10}
    \expandafter\ifx\csname url\endcsname\relax
      \def\url#1{\texttt{#1}}\fi
    \expandafter\ifx\csname urlprefix\endcsname\relax\def\urlprefix{URL }\fi
    \providecommand{\bibinfo}[2]{#2}
    \providecommand{\eprint}[2][]{\url{#2}}
    
    \bibitem{Crocker2000}
    \bibinfo{author}{Crocker, J.} \emph{et~al.}
    \newblock \bibinfo{title}{Two-point microrheology of inhomogeneous soft
      materials}.
    \newblock \emph{\bibinfo{journal}{Phys. Rev. Lett.}}
      \textbf{\bibinfo{volume}{85}}, \bibinfo{pages}{888--891}
    \newblock  (\bibinfo{year}{2000}).
    
    \bibitem{Lau2003}
    \bibinfo{author}{Lau, A.}, \bibinfo{author}{Hoffman, B.},
      \bibinfo{author}{Davies, A.}, \bibinfo{author}{Crocker, J.} \&
      \bibinfo{author}{Lubensky, T.}
    \newblock \bibinfo{title}{Microrheology, stress fluctuations, and active
      behavior of living cells}.
    \newblock \emph{\bibinfo{journal}{Phys. Rev. Lett.}}
    \textbf{\bibinfo{volume}{91}}, \bibinfo{pages}{198101-1--198101-4}
    \newblock  (\bibinfo{year}{2003}).
    
    \bibitem{Burghelea2004}
    \bibinfo{author}{Burghelea, T.}, \bibinfo{author}{Segre, E.},
      \bibinfo{author}{Bar-Joseph, I.}, \bibinfo{author}{Groisman, A.} \&
      \bibinfo{author}{Steinberg, V.}
    \newblock \bibinfo{title}{Chaotic flow and efficient mixing in a microchannel
      with a polymer solution}.
    \newblock \emph{\bibinfo{journal}{Phys. Rev. E}} \textbf{\bibinfo{volume}{69}},
      \bibinfo{pages}{066305}
    \newblock  (\bibinfo{year}{2004}).
    
    \bibitem{Gerashchenko2005}
    \bibinfo{author}{Gerashchenko, S.}, \bibinfo{author}{Chevallard, C.} \&
      \bibinfo{author}{Steinberg, V.}
    \newblock \bibinfo{title}{Single polymer dynamics: coil-stretch transition in a
      random flow}.
    \newblock \emph{\bibinfo{journal}{Europhys. Lett.}}
      \textbf{\bibinfo{volume}{71}}, \bibinfo{pages}{221--227}
    \newblock  (\bibinfo{year}{2005}).
    
    \bibitem{Gosse2002}
    \bibinfo{author}{Gosse, C.} \& \bibinfo{author}{Croquette, V.}
    \newblock \bibinfo{title}{Magnetic tweezers: Micromanipulation and force
      measurement at the molecular level}.
    \newblock \emph{\bibinfo{journal}{Biophys. J.}}
      \textbf{\bibinfo{volume}{82}}, \bibinfo{pages}{3314--3329}
    \newblock  (\bibinfo{year}{2002}).
    
    \bibitem{deMello2006}
    \bibinfo{author}{deMello, A.~J.}
    \newblock \bibinfo{title}{Control and detection of chemical reactions in
      microfluidic systems}.
    \newblock \emph{\bibinfo{journal}{Nature}} \textbf{\bibinfo{volume}{442}},
      \bibinfo{pages}{394--402}
    \newblock  (\bibinfo{year}{2006}).
    
    \bibitem{McMullen2010}
    \bibinfo{author}{McMullen, J.~P.} \& \bibinfo{author}{Jensen, K.~F.}
    \newblock \bibinfo{title}{Integrated microreactors for reaction automation: New
      approaches to reaction development}.
    \newblock \emph{\bibinfo{journal}{Annu. Rev. Anal. Chem.}}
      \textbf{\bibinfo{volume}{3}}, \bibinfo{pages}{19--42}
    \newblock  (\bibinfo{year}{2010}).
    
    \bibitem{Khandurina2000}
    \bibinfo{author}{Khandurina, J.} \emph{et~al.}
    \newblock \bibinfo{title}{Integrated system for rapid {PCR}-based {DNA}
      analysis in microfluidic devices}.
    \newblock \emph{\bibinfo{journal}{Anal. Chem.}}
      \textbf{\bibinfo{volume}{72}}, \bibinfo{pages}{2995--3000}
    \newblock  (\bibinfo{year}{2000}).
    
    \bibitem{Zhang2006}
    \bibinfo{author}{Zhang, C.}, \bibinfo{author}{Xu, J.}, \bibinfo{author}{Ma, W.}
      \& \bibinfo{author}{Zheng, W.}
    \newblock \bibinfo{title}{{PCR} microfluidic devices for {DNA} amplification}.
    \newblock \emph{\bibinfo{journal}{Biotechnol. Adv.}}
      \textbf{\bibinfo{volume}{24}}, \bibinfo{pages}{243--284}
    \newblock  (\bibinfo{year}{2006}).
    
    \bibitem{Sackmann2014}
    \bibinfo{author}{Sackmann, E.~K.}, \bibinfo{author}{Fulton, A.~L.} \&
      \bibinfo{author}{Beebe, D.~J.}
    \newblock \bibinfo{title}{The present and future role of microfluidics in
      biomedical research}.
    \newblock \emph{\bibinfo{journal}{Nature}} \textbf{\bibinfo{volume}{507}},
      \bibinfo{pages}{181--189}
    \newblock  (\bibinfo{year}{2014}).
    
    \bibitem{Bourgoin2006}
    \bibinfo{author}{Bourgoin, M.}, \bibinfo{author}{Ouellette, N.~T.},
      \bibinfo{author}{Xu, H.}, \bibinfo{author}{Berg, J.} \&
      \bibinfo{author}{Bodenschatz, E.}
    \newblock \bibinfo{title}{The role of pair dispersion in turbulent flow}.
    \newblock \emph{\bibinfo{journal}{Science}} \textbf{\bibinfo{volume}{311}},
      \bibinfo{pages}{835--838}
    \newblock  (\bibinfo{year}{2006}).

    \bibitem{Afik2014}
    \bibinfo{author}{Afik, E.} \& \bibinfo{author}{Steinberg, V.}
    \newblock \bibinfo{title}{Pair dispersion in a chaotic flow reveals the role of
      the memory of initial velocity}.
      \newblock \emph{\bibinfo{journal}{e-prints ArXiv:1502.02818v1}}
    \newblock  (\bibinfo{year}{2015}).
    \newblock \bibinfo{note}{Submitted}.
    
    \bibitem{Speidel2003}
    \bibinfo{author}{Speidel, M.}, \bibinfo{author}{Jon\'{a}\v{s}, A.} \&
      \bibinfo{author}{Florin, E.-L.}
    \newblock \bibinfo{title}{Three-dimensional tracking of fluorescent
      nanoparticles with subnanometer precision by use of off-focus imaging}.
    \newblock \emph{\bibinfo{journal}{Opt. Lett.}}
      \textbf{\bibinfo{volume}{28}}, \bibinfo{pages}{69--71}
    \newblock  (\bibinfo{year}{2003}).
   
    \bibitem{Ginkel2004}
    \bibinfo{author}{van Ginkel, M.}, \bibinfo{author}{Hendriks, C.~L.} \&
      \bibinfo{author}{van Vliet, L.}
    \newblock \bibinfo{title}{A short introduction to the radon and hough
      transforms and how they relate to each other}.
    \newblock \bibinfo{type}{Tech. Rep.} \bibinfo{number}{QI-2004-01 in the
      Quantitative Imaging Group Technical Report Series},
      \bibinfo{institution}{Delft University of Technology}
    \newblock  (\bibinfo{year}{2004}).
    
    \bibitem{Duda1972}
    \bibinfo{author}{Duda, R.~O.} \& \bibinfo{author}{Hart, P.~E.}
    \newblock \bibinfo{title}{Use of the hough transformation to detect lines and
      curves in pictures}.
    \newblock \emph{\bibinfo{journal}{Comm. ACM}}
      \textbf{\bibinfo{volume}{15}}, \bibinfo{pages}{11--15}
    \newblock  (\bibinfo{year}{1972}).
    
    \bibitem{Yuen1990}
    \bibinfo{author}{Yuen, H.}, \bibinfo{author}{Princen, J.},
      \bibinfo{author}{Illingworth, J.} \& \bibinfo{author}{Kittler, J.}
    \newblock \bibinfo{title}{Comparative study of hough transform methods for
      circle finding}.
    \newblock \emph{\bibinfo{journal}{Image and Vision Computing}}
      \textbf{\bibinfo{volume}{8}}, \bibinfo{pages}{71--77}
    \newblock  (\bibinfo{year}{1990}).
    
    \bibitem{Huang2012}
    \bibinfo{author}{Huang, Y.-H.}, \bibinfo{author}{Chung, K.-L.},
      \bibinfo{author}{Yang, W.-N.} \& \bibinfo{author}{Chiu, S.-H.}
    \newblock \bibinfo{title}{Efficient symmetry-based screening strategy to speed
      up randomized circle-detection}.
    \newblock \emph{\bibinfo{journal}{Pattern Recogn. Lett.}}
      \textbf{\bibinfo{volume}{33}}, \bibinfo{pages}{2071--2076}
    \newblock  (\bibinfo{year}{2012}).
    
    \bibitem{Lindeberg1999}
    \bibinfo{author}{Lindeberg, T.}
    \newblock \bibinfo{title}{Principles for automatic scale selection}.
    \newblock In \bibinfo{editor}{J{\"a}hne, B.}, \bibinfo{editor}{Hau{\ss}ecker,
      H.} \& \bibinfo{editor}{Gei{\ss}ler, P.} (eds.)
      \emph{\bibinfo{booktitle}{{H}andbook on {C}omputer {V}ision and
      {A}pplications}}, vol.~\bibinfo{volume}{2}, \bibinfo{pages}{239--274}
    \newblock  (\bibinfo{publisher}{Academic Press}, \bibinfo{year}{1999}).
    
    \bibitem{Kimme1975}
    \bibinfo{author}{Kimme, C.}, \bibinfo{author}{Ballard, D.} \&
      \bibinfo{author}{Sklansky, J.}
    \newblock \bibinfo{title}{Finding circles by an array of accumulators}.
    \newblock \emph{\bibinfo{journal}{Comm. ACM}}
      \textbf{\bibinfo{volume}{18}}, \bibinfo{pages}{120--122}
    \newblock  (\bibinfo{year}{1975}).
    
    %\bibitem{Afik2014a}
    %\bibinfo{author}{Afik, E.}
    %\newblock \bibinfo{title}{ridge directed ring detector}.
    %\newblock
    %  \bibinfo{howpublished}{\url{https://github.com/eldad-a/ridge-directed-ring-detector}}
    %\newblock  (\bibinfo{year}{2014}).
   
    \bibitem{Akinlar2013}
    \bibinfo{author}{Akinlar, C.} \& \bibinfo{author}{Topal, C.}
    \newblock \bibinfo{title}{{EDC}ircles: A real-time circle detector with a false
      detection control}.
    \newblock \emph{\bibinfo{journal}{Pattern Recognit.}}
      \textbf{\bibinfo{volume}{46}}, \bibinfo{pages}{725--740}
    \newblock  (\bibinfo{year}{2013}).
    
    \bibitem{Desolneux2007}
    \bibinfo{author}{Desolneux, A.}, \bibinfo{author}{Moisan, L.} \&
      \bibinfo{author}{Morel, J.-M.}
    \newblock \emph{\bibinfo{title}{From gestalt theory to image analysis: a
      probabilistic approach}}, vol.~\bibinfo{volume}{34} of
      \emph{\bibinfo{series}{0939-6047}}
    \newblock  (\bibinfo{publisher}{Springer}, \bibinfo{year}{2007}).
     
    \bibitem{OpenCV}
    \bibinfo{author}{Bradski, G.}
    \newblock \bibinfo{title}{The opencv library}.
    \newblock \emph{\bibinfo{journal}{Dr. Dobb's Journal of Software Tools}}
    \newblock  (\bibinfo{year}{2000}).
    
    \bibitem{Amini2012}
    \bibinfo{author}{Amini, H.}, \bibinfo{author}{Sollier, E.},
      \bibinfo{author}{Weaver, W.~M.} \& \bibinfo{author}{Di~Carlo, D.}
    \newblock \bibinfo{title}{Intrinsic particle-induced lateral transport in
      microchannels}.
    \newblock \emph{\bibinfo{journal}{Proc. Natl. Acad. Sci. U.S.A.}}
    \textbf{\bibinfo{volume}{109}}, \bibinfo{pages}{11593--11598}
    \newblock  (\bibinfo{year}{2012}).
    
    \bibitem{Brandenburg2007}
    \bibinfo{author}{Brandenburg, B.} \& \bibinfo{author}{Zhuang, X.}
    \newblock \bibinfo{title}{Virus trafficking -- learning from single-virus
      tracking}.
    \newblock \emph{\bibinfo{journal}{Nature Rev. Microbiol.}}
      \textbf{\bibinfo{volume}{5}}, \bibinfo{pages}{197--208}
    \newblock  (\bibinfo{year}{2007}).
    
    \bibitem{Friedl2000}
    \bibinfo{author}{Friedl, P.} \& \bibinfo{author}{Br\"{o}cker, E.-B.}
    \newblock \bibinfo{title}{The biology of cell locomotion within
      three-dimensional extracellular matrix}.
    \newblock \emph{\bibinfo{journal}{Cell. Mol. Life Sci.}}
      \textbf{\bibinfo{volume}{57}}, \bibinfo{pages}{41--64}
    \newblock  (\bibinfo{year}{2000}).
    
    \bibitem{Xie2011}
    \bibinfo{author}{Xie, L.}, \bibinfo{author}{Altindal, T.},
      \bibinfo{author}{Chattopadhyay, S.} \& \bibinfo{author}{Wu, X.-L.}
    \newblock \bibinfo{title}{Bacterial flagellum as a propeller and as a rudder
      for efficient chemotaxis}.
    \newblock \emph{\bibinfo{journal}{Proc. Natl. Acad. Sci. U.S.A.}}
    \textbf{\bibinfo{volume}{108}}, \bibinfo{pages}{2246--2251}
    \newblock  (\bibinfo{year}{2011}).
    
    \bibitem{Cython}
    \bibinfo{author}{Bradshaw, R.} %, \bibinfo{author}{Behnel, S.},
    %  \bibinfo{author}{Seljebotn, D.}, \bibinfo{author}{Ewing, G.}
       \emph{et~al.}
    \newblock \bibinfo{title}{The {C}ython compiler}.
    \newblock Available at: 
    \newblock \bibinfo{howpublished}{http://cython.org/}.
    \newblock \bibinfo{note}{(Accessed: 1st October 2014)}.
    
    \bibitem{SciPy}
    \bibinfo{author}{Jones, E.} %, \bibinfo{author}{Oliphant, T.},
     %  \bibinfo{author}{Peterson, P.}
        \emph{et~al.}
    \newblock \bibinfo{title}{{SciPy}: Open source scientific tools for {Python}}.
    \newblock  (\bibinfo{year}{2001--}).
    \newblock Available at: 
    \newblock \bibinfo{howpublished}{http://www.scipy.org/}
    \newblock \bibinfo{note}{(Accessed: 1st October 2014)}.
   
    \bibitem{Wahba1983}
    \bibinfo{author}{Wahba, G.}
    \newblock \bibinfo{title}{Bayesian ``confidence intervals'' for the
      cross-validated smoothing spline}.
    \newblock \emph{\bibinfo{journal}{J. R. Stat. Soc. Series B}} \textbf{\bibinfo{volume}{45}},
      \bibinfo{pages}{133--150}
    \newblock  (\bibinfo{year}{1983}).
    
    \bibitem{Kelley2011}
    \bibinfo{author}{Kelley, D.~H.} \& \bibinfo{author}{Ouellette, N.~T.}
    \newblock \bibinfo{title}{Using particle tracking to measure flow instabilities
      in an undergraduate laboratory experiment}.
    \newblock \emph{\bibinfo{journal}{Am. J. Phys.}} \textbf{\bibinfo{volume}{79}},
      \bibinfo{pages}{267}
    \newblock  (\bibinfo{year}{2011}).
    
    \bibitem{Wasserm2007}
    \bibinfo{author}{Wasserman, L.}
    \newblock \emph{\bibinfo{title}{All of Nonparametric Statistics (Springer Texts
      in Statistics)}}
    \newblock  (\bibinfo{publisher}{Springer}, \bibinfo{year}{2007}).
    
    \bibitem{Ahnert2007}
    \bibinfo{author}{Ahnert, K.} \& \bibinfo{author}{Abel, M.}
    \newblock \bibinfo{title}{Numerical differentiation of experimental data: local
      versus global methods}.
    \newblock \emph{\bibinfo{journal}{Comput. Phys. Commun.}}
      \textbf{\bibinfo{volume}{177}}, \bibinfo{pages}{764--774}
    \newblock  (\bibinfo{year}{2007}).
    
    \bibitem{Krakauer2014}
    \bibinfo{author}{Krakauer, N.~Y.} \& \bibinfo{author}{Fekete, B.~M.}
    \newblock \bibinfo{title}{Are climate model simulations useful for forecasting
      precipitation trends? hindcast and synthetic-data experiments}.
    \newblock \emph{\bibinfo{journal}{Environ. Res. Lett.}}
      \textbf{\bibinfo{volume}{9}}, \bibinfo{pages}{024009}
    \newblock  (\bibinfo{year}{2014}).
    
    \bibitem{Cheong2010}
    \bibinfo{author}{Cheong, F.~C.}, \bibinfo{author}{Krishnatreya, B.~J.} \&
      \bibinfo{author}{Grier, D.~G.}
    \newblock \bibinfo{title}{Strategies for three-dimensional particle tracking
      with holographic video microscopy}.
    \newblock \emph{\bibinfo{journal}{Opt. Express}}
      \textbf{\bibinfo{volume}{18}}, \bibinfo{pages}{13563--13573}
    \newblock  (\bibinfo{year}{2010}).
    
    \bibitem{Dixon2011}
    \bibinfo{author}{Dixon, L.}, \bibinfo{author}{Cheong, F.~C.} \&
      \bibinfo{author}{Grier, D.~G.}
    \newblock \bibinfo{title}{Holographic deconvolution microscopy for
      high-resolution particle tracking}.
    \newblock \emph{\bibinfo{journal}{Opt. Express}}
      \textbf{\bibinfo{volume}{19}}, \bibinfo{pages}{16410--16417}
    \newblock  (\bibinfo{year}{2011}).
    
    \bibitem{Kao1994}
    \bibinfo{author}{Kao, H.} \& \bibinfo{author}{Verkman, A.}
    \newblock \bibinfo{title}{Tracking of single fluorescent particles in three
      dimensions: use of cylindrical optics to encode particle position}.
    \newblock \emph{\bibinfo{journal}{Biophys. J.}}
      \textbf{\bibinfo{volume}{67}}, \bibinfo{pages}{1291--1300}
    \newblock  (\bibinfo{year}{1994}).
    
    \bibitem{Huang2008}
    \bibinfo{author}{Huang, B.}, \bibinfo{author}{Wang, W.},
      \bibinfo{author}{Bates, M.} \& \bibinfo{author}{Zhuang, X.}
    \newblock \bibinfo{title}{Three-dimensional super-resolution imaging by
      stochastic optical reconstruction microscopy}.
    \newblock \emph{\bibinfo{journal}{Science}} \textbf{\bibinfo{volume}{319}},
      \bibinfo{pages}{810--813}
    \newblock  (\bibinfo{year}{2008}).
    
    \bibitem{Babcock2012}
    \bibinfo{author}{Babcock, H.}, \bibinfo{author}{Sigal, Y.~M.} \&
      \bibinfo{author}{Zhuang, X.}
    \newblock \bibinfo{title}{A high-density {3D} localization algorithm for
      stochastic optical reconstruction microscopy}.
    \newblock \emph{\bibinfo{journal}{Opt. Nanoscopy}}
      \textbf{\bibinfo{volume}{1}}, \bibinfo{pages}{6}
    \newblock  (\bibinfo{year}{2012}).
 
\end{thebibliography}
%\input{EA_Ridge_directed_ring_detection.bbl}

\section*{Acknowledgements}
    I thank P. Reisman for introducing me to the Hough transfrom, O. Schwatz for
    helpful discussions regarding the optical setup, A. Frishman
    and S. van der Walt for their useful comments on the manuscript. 
    Special thanks go to Y. Kaplan and T. Afik for their help in robustness assessment,
    and to V. Steinberg and M. Feldman for helpful discussions of this work and
    its presentation.
    This work is supported by grants from the German-Israel Foundation (GIF)
    and the Lower Saxony Ministry of Science and Culture Cooperation
    (Germany).

\section*{Additional Information}
%\paragraph*{Supplementary information} accompanies this paper at
%\ref{http://www.nature.com/article-assets/npg/srep/2015/150902/srep13584/extref/srep13584-s1.pdf} 
% \iffalse http://www.nature.com/srep \fi

\paragraph*{Competing financial interests:}
The author declares no competing financial interests.

\paragraph*{How to cite this article:}
Afik, E. 
Robust and highly performant ring detection algorithm for 3d particle
tracking using 2d microscope imaging.
Sci. Rep. 5, 13584; 
\href{http://www.nature.com/articles/srep13584}{doi: 10.1038/srep13584} (2015)

\bigskip
\noindent This work is licensed under a Creative Commons Attribution 4.0 International License. The
images or other third party material in this article are included in the article’s Creative Com-
mons license, unless indicated otherwise in the credit line; if the material is not included under the
Creative Commons license, users will need to obtain permission from the license holder to reproduce
the material. To view a copy of this license, visit
\url{http://creativecommons.org/licenses/by/4.0/}

%%%%%%%%%%%%%%%%%%%%%%%%%%%%%%%%%%%%%%%%%%%%%%%%%%%%%%%%%%%%%%%%%%%%%%%%%%%%%%%%%%%
%%% TODO SciRep Fig Submission

\newpage

\ifSciRepSubmission
\begin{figure*}[<+htpb+>]
    \caption{\capheader{Snapshots from the experiment and a demonstration of the algorithm
    robustness}
        \subcap{a} typical image complexity is exemplified in an unprocessed
        sub-frame consisting of 1/9 part of the full frame, corresponding to
        lateral dimension of \SI{215}{\um}$\times$\SI{315}{\um}. The axial range available
        for the particles is \SI{140}{\um}.
        \subcap{b} the corresponding analysis result; in red are the radii in
        pixels units.
        \subcap{c} \& \subcap{d} time sequences of sub-frames (\SI{400}{\ms} each). 
        Red coloured particles in \subcap{c} demonstrate pair dispersion, in
        which the algorithm is required to resolve rings with similar
        parameters.
        The yellow particle in \subcap{d} shows radius change corresponding to a
        downwards translation.
        Each sub-frame in \subcap{c} \& \subcap{d} images a box which lateral
        dimensions is \SI{190}{\um}$\times$\SI{270}{\um}.
        }\label{fig:video_snapshots}
\end{figure*}
\fi

\ifSciRepSubmission
\begin{figure*}[<+htpb+>]
   \caption{\capheader{Algorithm outline}
        \subcap{a} raw sub-image containing two fluorescent particles; note
        that the inner rings of each particle are thinner than the outer most
        one. This scale separation admits suppression of all but the outer most
        ring via Gaussian smoothing (to ease visualisation the contrast was
        enhanced in the images on the expense of the central peak of the
        diffraction pattern);
        \subcap{b} ridge detection: the ridges are defined using a differential
        geometric descriptor and shown here as arrows representing $\curvdir$,
        the principal direction, corresponding to $\curv$, the least principal
        curvature, which is plotted in the background. The arrows originate
        from the ridge pixel. Note that the inner rings are successfully
        suppressed based on the scale separation. To ease visualisation every
        second detected ridge is omitted;
        \subcap{c} circle Hough transform: directed ridges $\to$ circle
        parameter space;
        \subcap{d} local maxima detection: radius dependent smoothing of the
        parameter space as well as normalisation by 1/r and thresholding
        greatly emphasise the local maxima representing the rings in the image;
        \subcap{e} sub-pixel accuracy: based on the detected rings, annulus
        masks (blue and green annuli in the figure) allow classification of
        ridge pixels (red points) and sub-pixel accuracy is achieved via circle
        fitting. Note the discarded directed ridges of the central peak (in
        \subcap{b}) as they do not belong to any local maxima in the processed
        circle parameter space \subcap{d};
        \subcap{f} the output: best fit circle for the ridge pixels of the
        outer-most ring of each particle.
        } \label{fig:algorithm_outline}
\end{figure*}
\fi

%%%%%%%%%%%%%%%%%%%%%%%%%%%%%%%
%%%          SI 
\ifCompileSI
\clearpage

\section*{{\huge\SInfo~for:}}

\section*{\LARGE{Robust and highly performant ring detection algorithm 
for 3d particle tracking using 2d microscope imaging}}

\renewcommand{\figurename}{Supplementary Figure}
\makeatletter 
\renewcommand{\thefigure}{S\@arabic\c@figure}
\setcounter{figure}{0}

\author{\large{Eldad Afik \\}
  \normalsize{Department of Physics of Complex Systems,}\\
  \normalsize{Weizmann Institute of Science,}\\
  \normalsize{Rehovot 76100, Israel}\\
  \\
  \normalsize{email: \href{mailto:eldada.afik@weizmann.ac.il}{eldad.afik@weizmann.ac.il}}
}
\date{}

\maketitle

\renewcommand{\abstractname}{\vspace{-\baselineskip}} %% skip the ``abstract'' header
\begin{abstract}
  To keep the main presentation succinct many details of the proposed algorithm and
  the application for particle tracking under the microscope were left out of the main text. 
  These can be found below.
  The Supplementary Information opens with a more detailed and technical presentation of the
  algorithm\iffalse\autoref{sec:detailed_algorithm}\fi, followed by the discussion of the
  application of the proposed algorithm for particle tracking, in particular in
  the light of the available alternatives%
  \iffalse\autoref{sec:application_for_tracking}\fi; 
  the closing section contains the supporting figures%
  \iffalse\autoref{sec:supporting_figures}\fi:  
  \begin{inparaenum}[(i)]
    \item  an empirical calibration curve of the ring radius to the out-of-focus
      distance is shown in
      \SItem~%
      \ifCompileSI
        \autoref{fig:rad2z}%
      \else
        Fig.~S1%
      \fi; the error bars provide an estimate of the precision of the
      proposed method resulting from the combination of the algorithm with the
      optical system together; 
   \item \SItem~%
     \ifCompileSI
       \autoref{fig:2d->3d}%
     \else
       Fig.~S2%
     \fi~ 
     shows a sample of 3d particle trajectories reconstructed based on the
     proposed method; and finally
   \item examples from the comparative assessment of the algorithm robustness
     referred to in the manuscript and described in the \refMethods~can be found in 
     \SItem~%
      \ifCompileSI
        \autoref{fig:EDCircles_comparison}%
      \else
        Fig.~S3%
      \fi.
  \end{inparaenum}
\end{abstract}

{\section*{Detailed algorithm}\label{sec:detailed_algorithm}}

As mentioned in the main text, the standard circle Hough transform is often
avoided not only for its challenging local maximum detection in noisy 3d space
but for its heavy memory requirements as well.
The standard circle Hough transform requires a 3-dimensional array of
accumulators. The coordinates of each array element are the parameters of a
candidate circle. The value of the accumulator at these coordinates indicates
how well this circle is represented in the image.
Code optimisation for high-performance and small memory footprint is achieved
following this scheme:

\begin{enumerate}
    \item \emph{Image pre-processing step:}
        the image is smoothed using a Gaussian convolution and the smoothed
        image spatial derivatives are calculated using a 5$\times$5
        2\textsuperscript{nd} order Sobel operator \cite{OpenCV}. Using
        these derivatives, the local least principal curvature $\curv$ is
        estimated as the smaller eigen-value of the Hessian matrix.
    \item \emph{One-pass ridge detection and votes collection:} 
        for each pixel in $\curv$ which is smaller than a pre-defined curvature
        threshold (the latter is no greater than zero), the corresponding
        $\curvdir$ is calculated. If this pixel is found to be a local
        minimum along the direction of $\curvdir$, its coordinates are
        recorded in the ridge container. At this stage, its votes are collected
        as well, that is, the potential circles parameters to which it may
        belong. 
    \item \emph{Sort the votes stack according to the radii:}
        this allows performing the parameter space incrementing procedure
        equi-radius level by level. In order to achieve higher performance, the
        votes are further sorted by the row index and then by the column index
        exploiting the numpy/cython strided direct data access
        \cite{Cython,SciPy}. For this reason each circle parameter triple is
        represented as an integer using a bijection.
    \item \emph{Circle parameter space population and local maximum detection
        via radius-dependent smoothing and normalisation:}
        This is done using two arrays representing a sub-space of the full
        circle parameter space. Each consists of 3 consecutive
        equi-radius levels; the first for the raw accumulators sub-space, the
        second for the smoothed and normalised one, where local maxima are to be
        searched for. There are two votes thresholding steps: an integer
        threshold for the raw accumulators and a floating point
        threshold, a fraction of $2\pi$, for the smoothed and normalised array
        elements.
        In describing the procedure it is assumed that the $r-2$ and $r-1$
        levels have already been populated in both sub-space triples and
        the $r$ levels are blank, i.e. all zeros. As long as the votes drawn
        from the votes stack point to the same radius, the corresponding
        radius-level is populated by incrementing the indicated accumulator. Recall
        that the votes are fully sorted hence all votes pointing to a certain
        voxel will come out from the stack in a row. Every time a new circle
        parameter triple is encountered, its coordinates are recorded as
        modified. In case the previously incremented voxel has surpassed the
        1\textsuperscript{st} votes threshold its coordinates are recorded as a
        hotspot -- a circle candidate.
        Once there are no more votes for this $r$-level, it is mapped to the
        second subspace: for each hotspot voxel a spatial average is
        calculated, weighted by a Gaussian function, which width is linearly
        dependent on the radius; the value of the average is then normalised by
        $1/r$.
        After mapping all the hostpots of the current $r$-level, a local
        maximum is searched for among the hotspots of the $(r-1)$-level
        which pass the 2\textsuperscript{nd} votes threshold. This is done
        using a nearest neighbours comparison within a $3\times3\times3$
        voxels box. Array elements which are local maximum and exceed the
        threshold are registered as rings.
        Once all hotspots have been processed, all modifications to the
        $(r-2)$-level are undone as its data are no longer needed. By this it
        is made ready to be regarded as the next $r$-level and a cyclic
        permutation among the levels takes place. In practice, this
        is performed by accessing the equi-radius levels using the modulo
        operation -- the radius indices are calculated using $r\pmod 3$.
    \item \emph{Sub-pixeling via circle fit:}
        the detected ridge coordinates are subjected to a circle fit via the
        non-exclusive classification induced by the results of the directed
        circle Hough transform. The coordinates in the ridge container are
        clustered based on annuli masks dictated by the detected rings and
        sub-pixel accuracy of the rings parameters is achieved.
\end{enumerate}

\subsection*{Additional notes}
\begin{itemize}
    \item The ridge detection can be used to achieve a compressed
        representation of the features in the image. This can be done by
        storing a hash table associating ridge coordinates as keys with their
        corresponding $\curvdir$ as values.
    \item The algorithm is not restricted to directed ridges as it can be
        replaced by directed edges in case these are better descriptors of the
        features in the image. This is achieved by replacing the Hessian by the
        Gradient. In this case, the gradient magnitude replacing $\curv$ has to
        be a local maximum along the gradient direction.
    \item To reduce false detection, the radii range is extended such that the
        Hough transform is over the range $\left[ r_{min}-1,r_{max}+1 \right]$,
        but local maximum detection are searched for within the original
        range.
    \item In case additional performance per processing unit is required, one
        could use a lower resolution in discretising the circle parameter
        space. Measuring the effect of this on the accuracy is left for future
        work.
    \item Using several colours, the method should be, in principle, extendible
        to even higher particle densities.
\end{itemize}

%\onecolumn
%\clearpage

\clearpage
%\vspace{-1cm}

\section*{Application of the proposed algorithm for particle tracking and
  discussion of alternative methods}\label{sec:application_for_tracking}

When tracking small light emitting objects, such as fluorescent particles under
the microscope, the appearance of rings is often a sign of the object going out
of focus. Normally this results in the loss of the tracked
object, which is thereafter considered as a hindering background source. 
However, these rings carry information of the 3-dimensional position of the
particle.
This has been used for localising a single light scattering magnetic bead
based on matching the radial intensity profile to an empirical set of reference
images \cite{Gosse2002}. An axial range of \SI{10}{\um} was demonstrated and a 
temporal resolution of \SI{25}{\Hz} was achieved using the knowledge of the 
particle's previous position.
In fact, for fluorescent particles the radius of the most visible
ring of each particle precisely indicates its axial position -- the radius follows a
simple scaling with the particle distance from the focal plane (see
\SItem~%
\ifCompileSI
  \autoref{fig:rad2z}%
\else
  Fig.~S1%
\fi). % \autoref{sec:supporting_figures}).
A similar approach was recently described in \cite{Speidel2003}, where the
measurements were, once again, limited to a single particle in the observation
volume, with an axial range of \SI{3}{\um} and temporal resolution of
\SI{10}{\Hz}.

In comparison with other existing methods for 3d particle tracking, the method
presented here is advantageous when it comes to long measurements, temporal
resolution and concurrency, as well as real-time applications.
The confocal scanning microscope requires scanning the volume of interest.
Therefore it is slower and cannot yet provide instantaneous information of the
whole volume. 
Unlike Holographic microscopy \cite{Cheong2010,Dixon2011}, the proposed method
does not pose long and heavy computational demands which is restrictive
for real-time applications or when large datasets are required for statistics.

One could expect the optical method discussed here to produce patterns which
are symmetric about the focal plane. When this applies, it may result in an
ambiguity with respect to whether the particle is above or below focus. Our
optical arrangement (see the \refMethods) shows clear diffraction rings
only on one side.
Furthermore, as particles approach focus, the radius of the outer-most ring
becomes too small to resolve. For these reasons the focal plane is placed
outside the volume of interest (as reflected in the \SItem~%
\ifCompileSI
  \autoref{fig:rad2z}%
\else
  Fig.~S1%
\fi). 
%\autoref{sec:supporting_figures}). 
Optical astigmatism offers a mean for discriminating between the two sides of
the optical axis \cite{Kao1994,Huang2008,Babcock2012}. The introduction of a
cylindrical lens results in the deformation of a circular spot into an
ellipsoidal one as a fluorescent particle goes further away from focus, with
the ellipse major axis of a particle above focus aligned perpendicular to a one
below. In Ref.~\cite{Kao1994} the axial range was limited to a couple of microns
above and below focus; in Refs.~\cite{Huang2008,Babcock2012} it was restricted to
less than a micron. Within these ranges the tracer image can be approximated by
an elliptical gaussian pattern. However, extending the range generates
elliptical rings as well; see Figure~1~in~Ref.~\cite{Kao1994}. This requires dealing with two
species of patterns, spots and rings. Moreover, deforming circular rings into
elliptical ones, the dimensionality of the parameter space increases, and so
does the technical complexity of the image analysis.
Therefore the advantage of the stronger signal, by working closer to focus on
both its sides, is expected to have a heavy computational cost once the range
is extended such that diffraction rings appear as well. 
The method presented here requires working away from focus. Rings
visibility decreases as the fluorescence signal spreads over a larger area,
thus setting the lower bound for the exposure time.
Nevertheless, I have found that the fluorescence signal-to-noise ratio allowed
tracking particles moving chaotically at speeds exceeding
\SI[per-mode=symbol]{400}{\um\per\s}.

\section*{Supplementary figures}\label{sec:supporting_figures}%
%\subsection{Calibration curve}
\begin{figure}[h!p]
    \centering{
    \tikzLabel{a}{%
      \includegraphics[width=.67\textwidth,trim=0 0 0 8mm,clip]{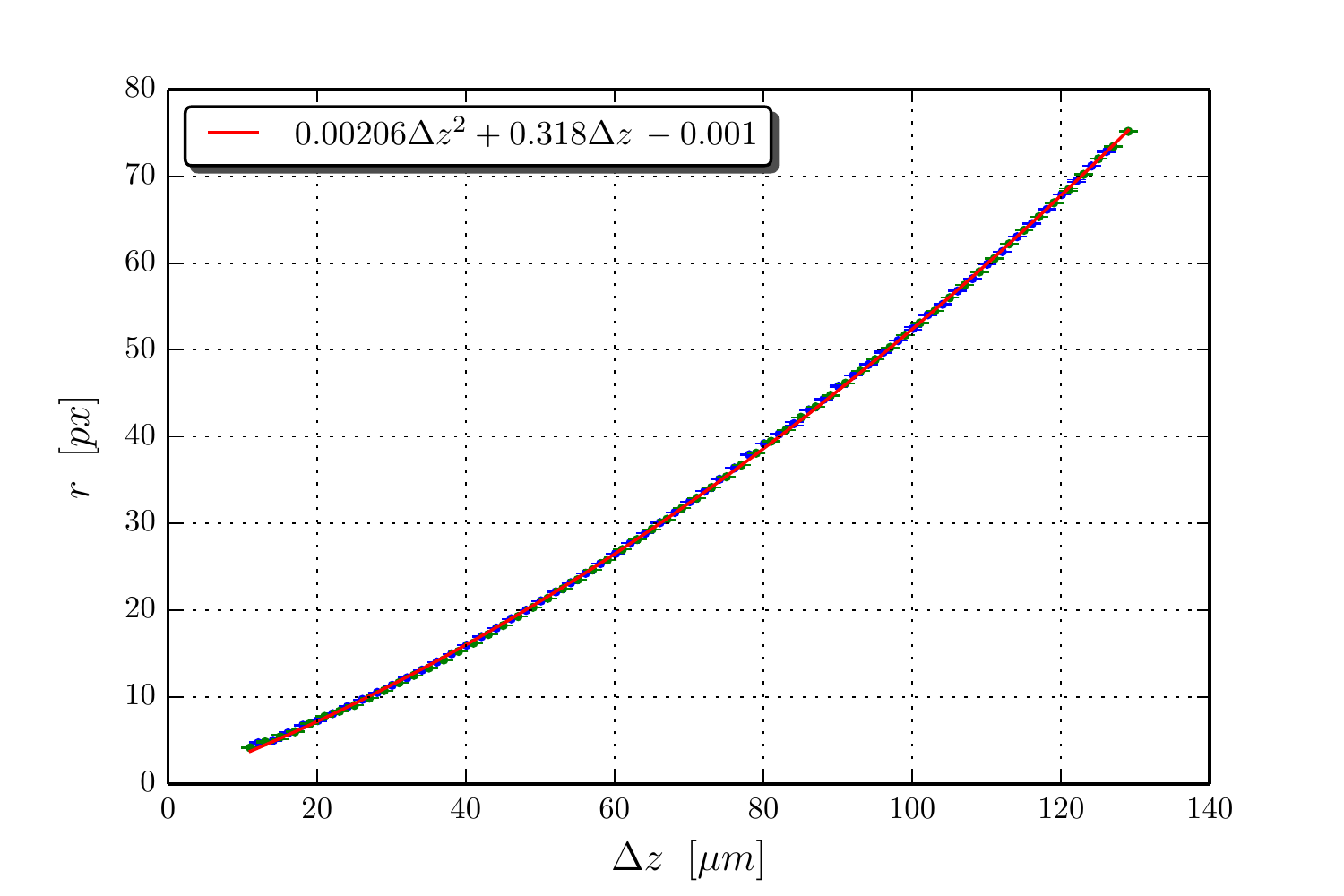}
    }
    \tikzLabel{b}{%
      \includegraphics[width=.67\textwidth,trim=0 0 0 8mm,clip]{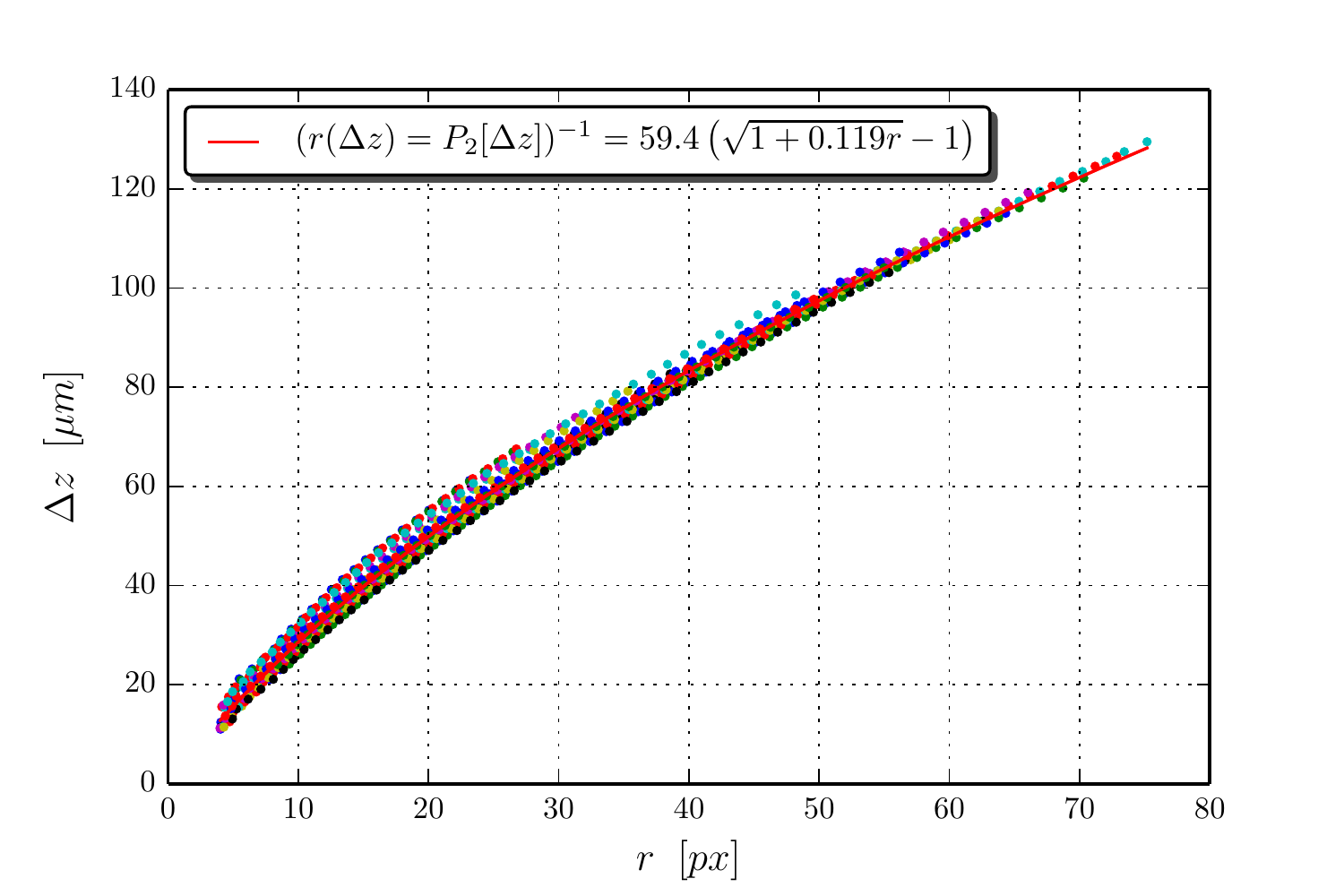}
    }}
    \caption{\capheader{Calibration curve}
      \subcap{a} The empirical relation between the outer-most ring radius and the
      out-of-focus distance $\Delta z$; the latter measures the translation of
      the objective from the position at which a particle would be in focus.
      The plot shows the data for two different fluorescent particles (denoted
      by blue and green in the plot).
      The quadratic polynomial fit provides an approximation for the 
      $r(\Delta z)$ relation.
      The data was acquired by scanning through the vertical axis of the
      observation volume by objective translation steps of \SI{2}{\um} (see
      Experimental details in the \refMethods).
      Each data point is an average of the measured radius over 210
      frames spanning \SI{3}{\second}, taken while the objective is stationary.
      Error bars reflect the standard-deviation; the median
      standard-deviation of the presented datasets is \SI{0.03}{\pixel} and the
      maximal is \SI{0.27}{\pixel}.
      \newline
      \subcap{b} The conversion function $r^{-1}(r)$ was obtained by the
      inversion of the quadratic polynomial fit, based on 25 tracers dispersed in the
      observation volume (see Experimental details in the \refMethods).
      The resulting root-mean-squared-error  
      $\sqrt{\langle \left( \Delta z - r^{-1}(r) \right)^2 \rangle}
      =$~\SI{1.97}{\um}, and the maximal measured absolute error is
      \SI{5.35}{\um}; these estimate the uncertainty due to the calibration
      procedure followed here.    
      Finally, the out-of-focus distance of the objective $\Delta z$ needs to
      be converted to a physical distance via multiplication by the refractive
      indices ratio, 1.58 in this case. Thus the observed axial range exceeds
      \SI{180}{\um}.
      }\label{fig:rad2z}
\end{figure}

%\subsection{2d out-of-focus image $\to$ trajectories}%
\begin{figure}[!hp]
    \begin{center}
      \begin{tikzpicture}[font=\sffamily]%\bfseries]%Helvetika] 
        \node[anchor=south west,inner sep=0] (image) at (0,0)%
        {\includegraphics[width=0.8\linewidth]{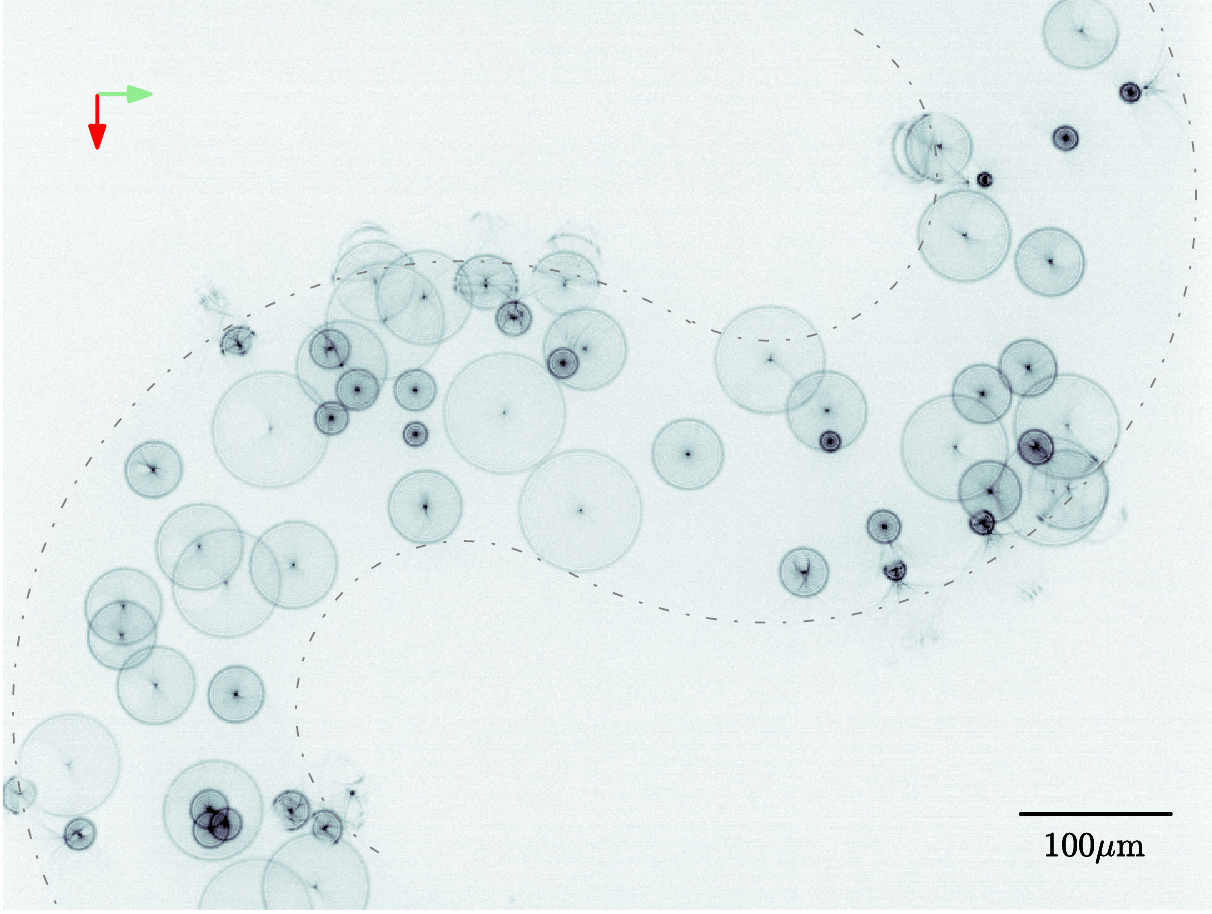}};
            \begin{scope}[x={(image.south east)},y={(image.north west)}]
                \draw (2.5mm,0.96) node {a};
            \end{scope}
        \end{tikzpicture}
    
        \tikzLabel{b}{%
            \includegraphics[width=0.8\linewidth]{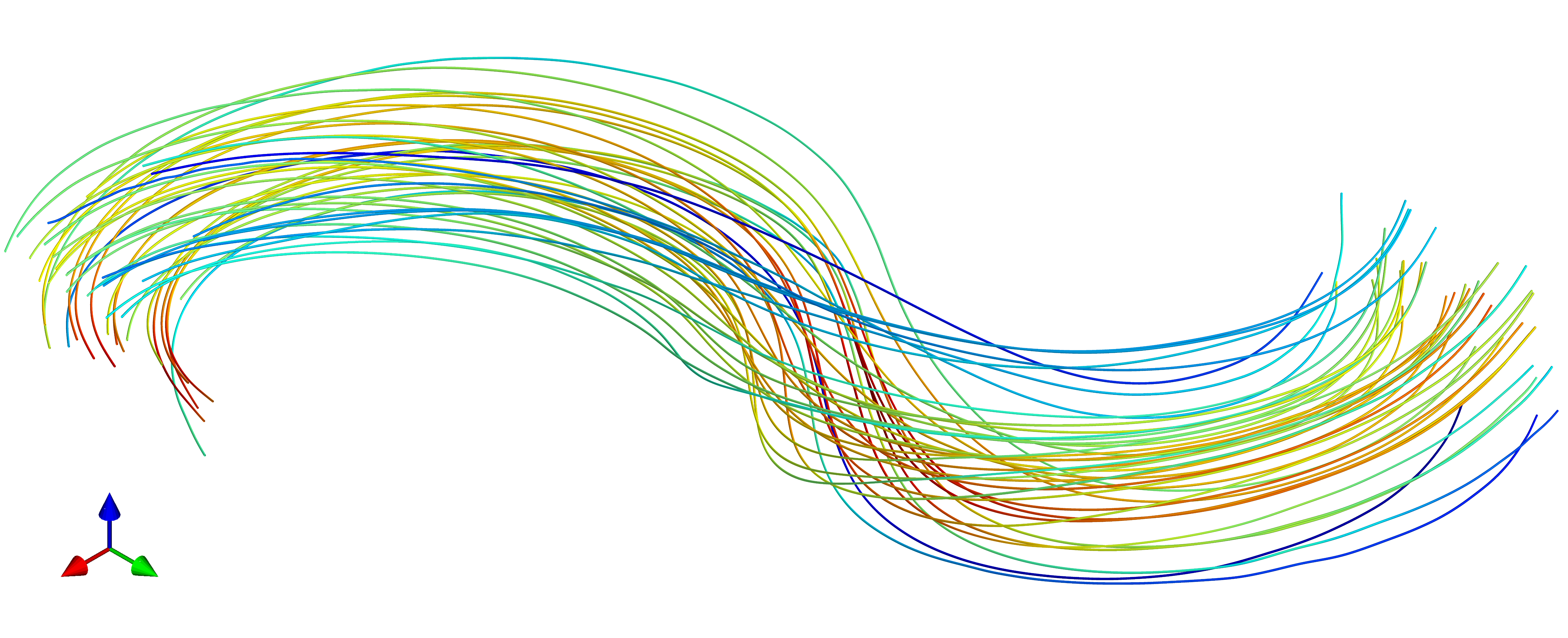}}
    \end{center}
    \caption{\capheader{2d out-of-focus images $\to$ 3d trajectories}
    \subcap{a} A single raw full 2d frame imaging a volume of 
        \SI{810}{\um}$\times$\SI{610}{\um}$\times$\SI{140}{\um}, including 60
        tracers in one period of the curvilinear tube (which boundaries are
        denoted by the broken grey line).
        Smaller rings result from tracers being closer to the focal plane, which
        is positioned above the tube.
        \subcap{b} A sub-sample of 40 trajectories reconstructed from a
        time-lapse sequence of such frames, which spans 12 seconds of data
        acquisition. The colour coding indicates the speed ranging from
        \SI[per-mode=symbol]{80}{\um\per\second} (blue) to
        \SI[per-mode=symbol]{400}{\um\per\second} (red); the mean flow is rightwards. 
        Corresponding axes are denoted by colours. 
        The isometric view of the bottom panel can be obtained by three
        rotations, starting from the orientation of the upper panel:
        \SI{-90}{\degree} about the green axis, \SI{-45}{\degree} about the blue
        axis, and approximately \SI{35}{\degree} about the new horizontal axis.
      }\label{fig:2d->3d}
\end{figure}

%\subsection{Comparative assessment of the algorithm robustness}
\begin{figure}[!hp]
    \begin{minipage}[b]{0.65\textwidth}
      %% http://tex.stackexchange.com/questions/222117/multiple-panel-figure-with-caption-in-one-of-the-panels
    \newlength\EAvsEDCsize
    \setlength\EAvsEDCsize{\textwidth}
    \centering
      \tikzLabel{a}{
          \includegraphics[width=\EAvsEDCsize,trim=0 7mm 0 7mm,clip]{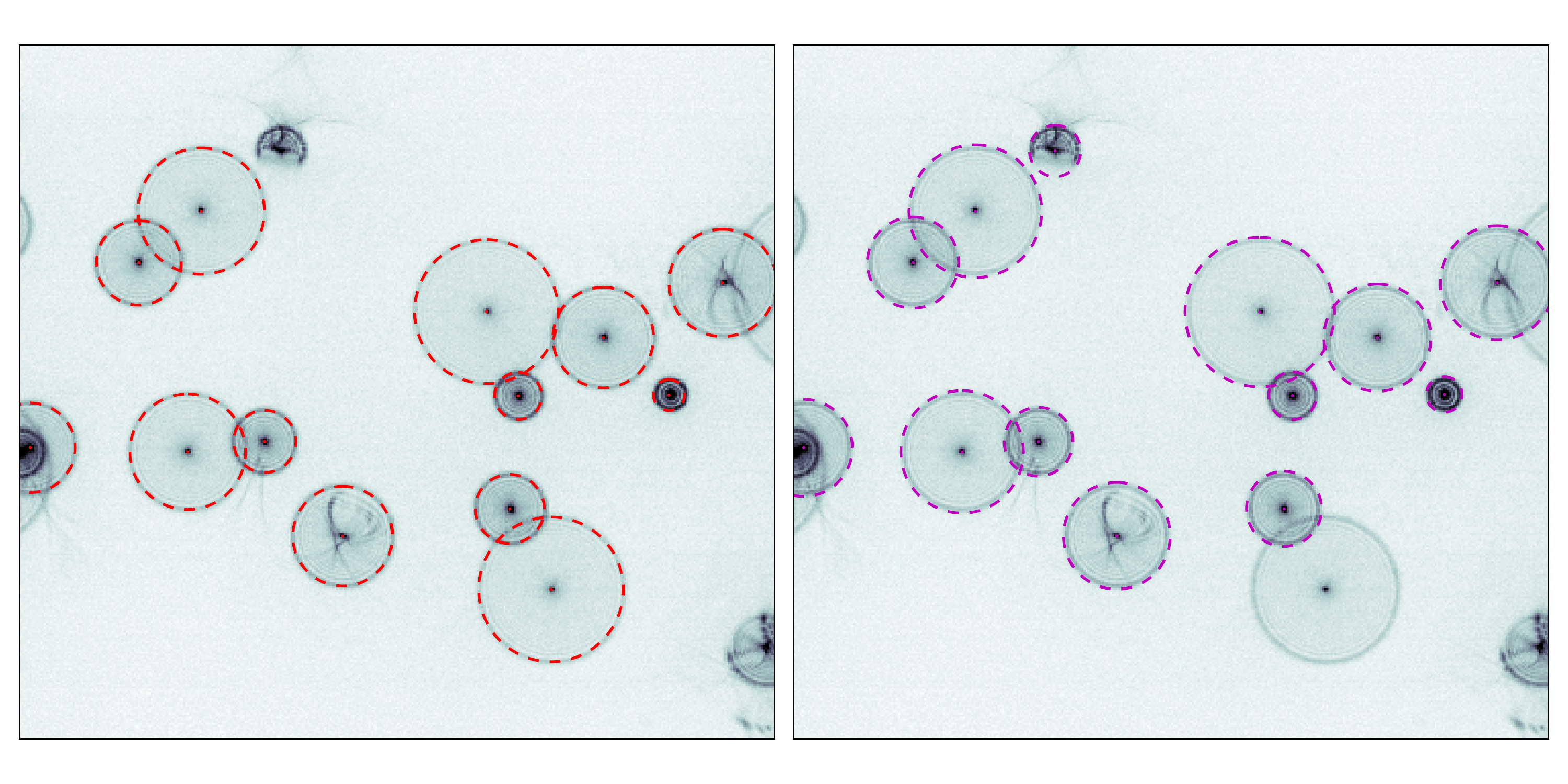}
        }\vspace{-2mm}

    \centering
      \tikzLabel{b}{
          \includegraphics[width=\EAvsEDCsize,trim=0 7mm 0 7mm,clip]{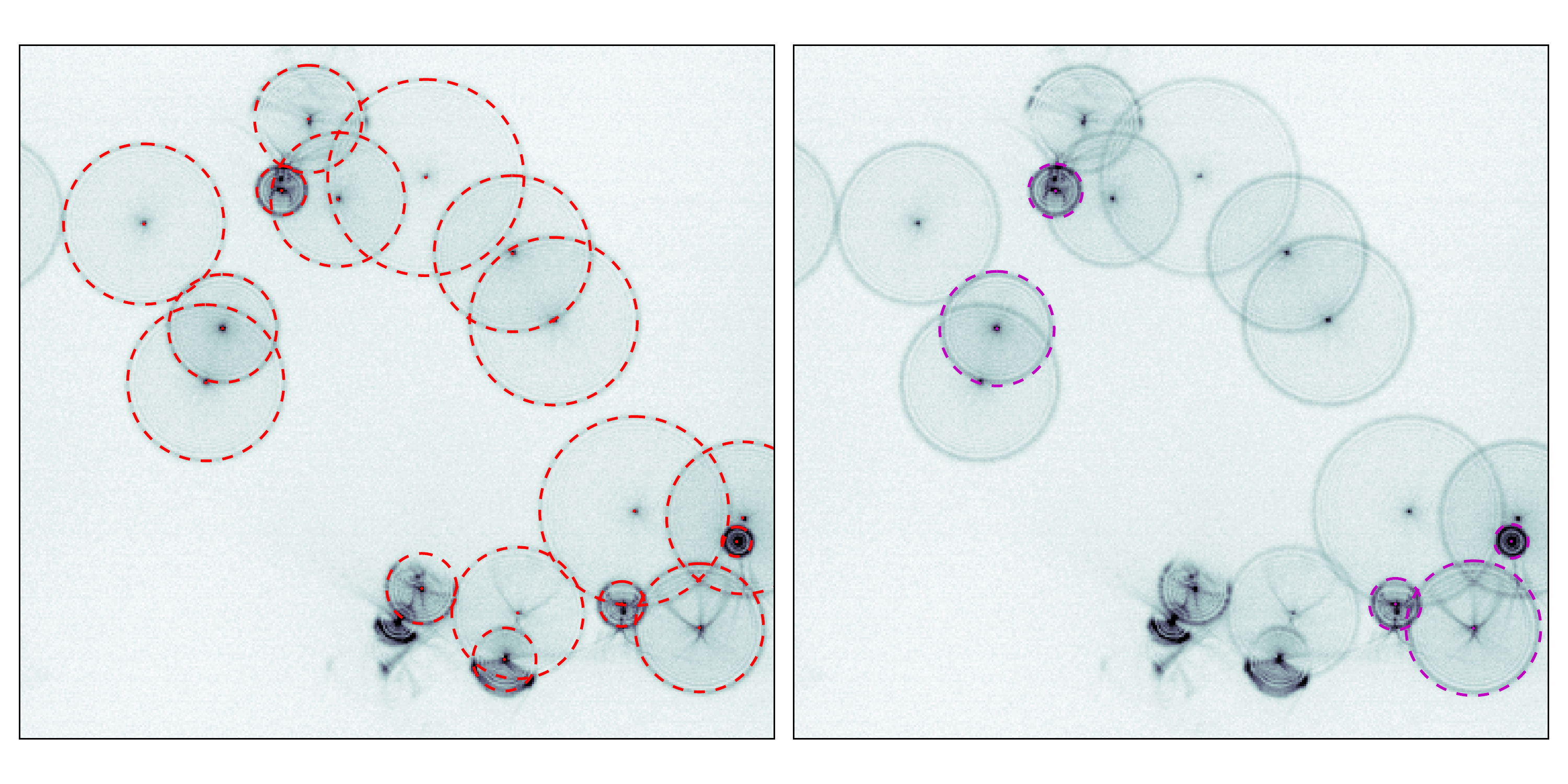}  
        }
    \centering
      \tikzLabel{c}{
          \includegraphics[width=\EAvsEDCsize,trim=0 7mm 0 7mm,clip]{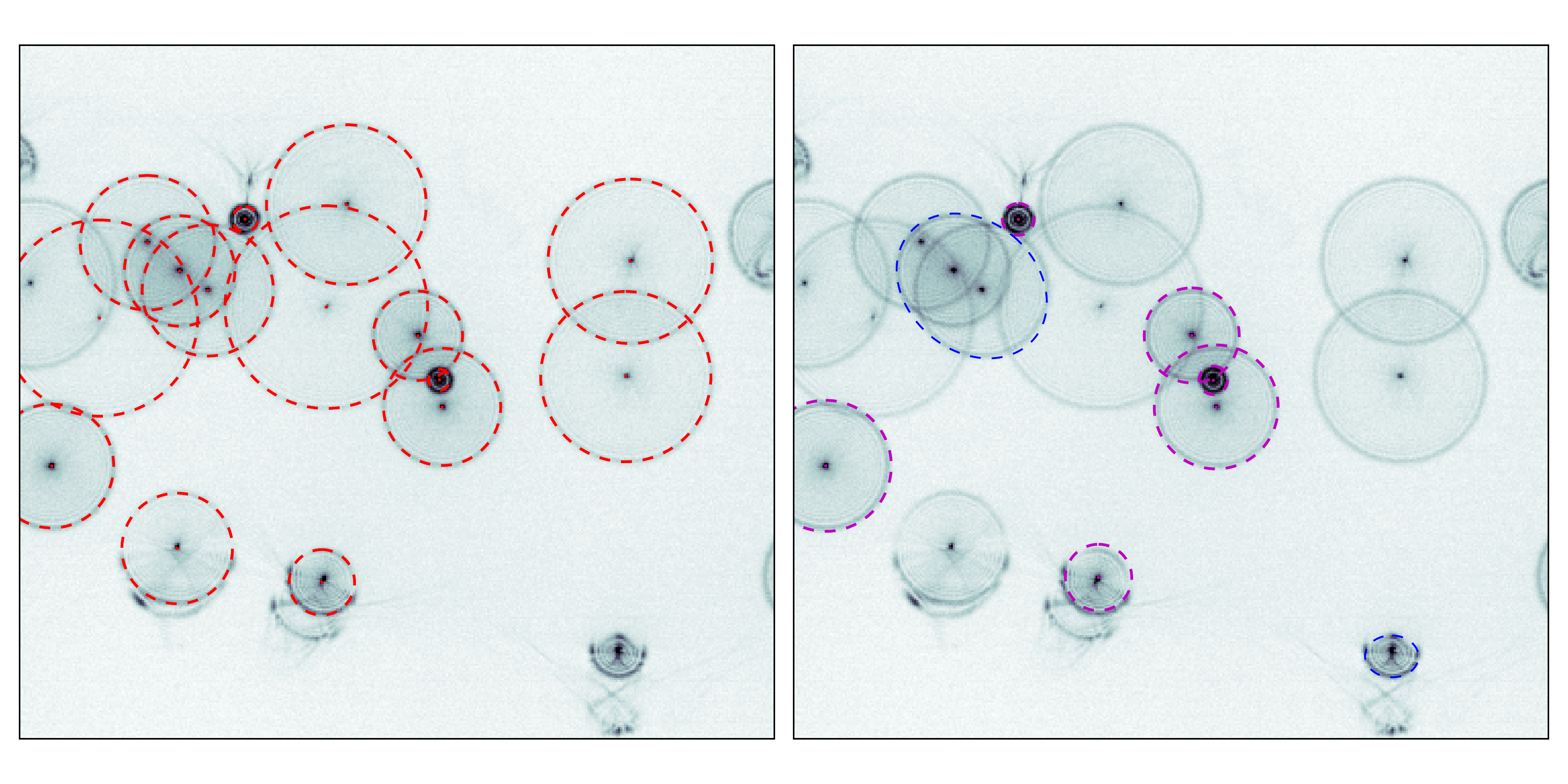}  
        }
    \centering
      \tikzLabel{d}{
          \includegraphics[width=\EAvsEDCsize,trim=0 7mm 0 7mm,clip]{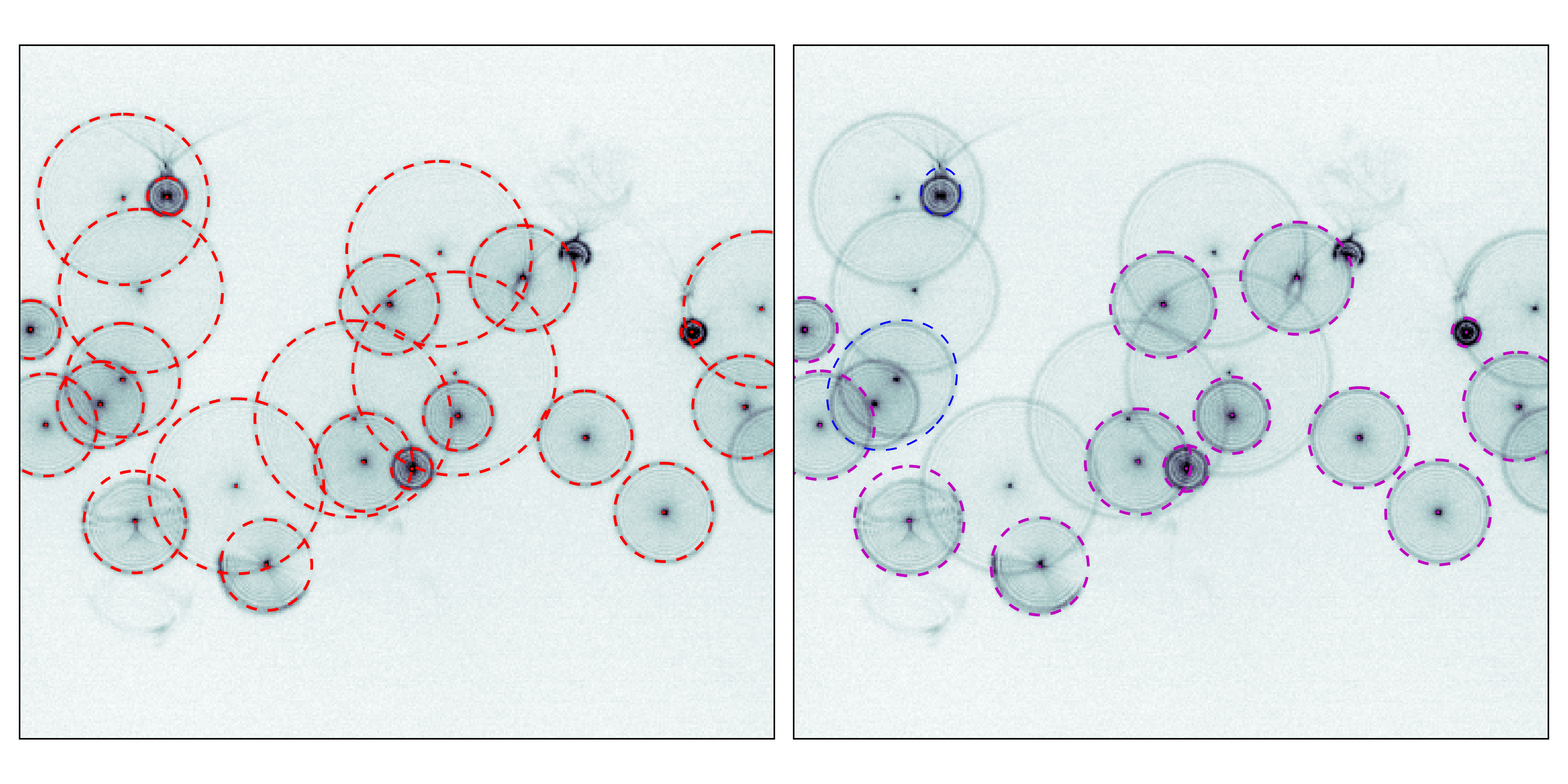}
        }
    \end{minipage}
    \begin{minipage}[b]{0.35\textwidth}
    \caption{\capheader{Examples from the comparative assessment of the algorithm
    robustness}
    Output examples of the proposed algorithm can be found on the left column 
    (reported rings are shown in dashed red) to be compared with those of EDCircles \cite{Akinlar2013} 
    on the right column (circles in purple, ellipses in blue);  
    further details of the test and results can be found in the \refMethods.
    In 3.3\% of the 151 tested images the alternative algorithm showed comparable
    results to those of the new one, that is both exhibited the same detection
    and error rates -- an example is shown in \subcap{a}; 
    in all other examined images the proposed method outperformed its
    competitor -- examples are shown in \subcap{b}, \subcap{c} \& \subcap{d}.
    The new algorithm resolves rings even when the signal-to-noise ratio is
    limiting for the opponent -- this is typical for tracers which are farther
    out-of-focus, hence their rings are larger and fainter. 
    Overlapping rings of similar radii are another challenge resolved by the new
    algorithm; in contrast these are often missed or merged into ellipses by
    the opponent -- see \subcap{c} \& \subcap{d}; similarly, optical artefacts
    often result in errors for the alternative method, in contrast to the new proposed one
    -- e.g. see small ring reported by the opponent to be an ellipse in \subcap{d}.
          }\label{fig:EDCircles_comparison}
     \end{minipage}
\end{figure}

\fi %% main ifCompileSI ?

\end{document}